%% file: main.tex
\documentclass[sigplan,twocolumn]{acmart}
\input{package}

\newboolean{publicversion}
\setboolean{publicversion}{true}

\date{}

\input{macros}

\input{0-abstract}

\begin{document}
\acmYear{2026}\copyrightyear{2026}
\setcopyright{cc}
\setcctype[4.0]{by}
\acmConference[EUROSYS '26]{European Conference on Computer Systems}{April 27--30, 2026}{Edinburgh, Scotland Uk}
\acmBooktitle{European Conference on Computer Systems (EUROSYS '26), April 27--30, 2026, Edinburgh, Scotland Uk}
\acmDOI{10.1145/3767295.3769366}
\acmISBN{979-8-4007-2212-7/26/04}

\title{\sysname: Optimizing Deep Learning Training Workloads using GPU Runtime Emulation}

\author{
Srihas~Yarlagadda$^{\dag*\text{1}}$\enskip
Amey~Agrawal$^{\text{*1}}$\enskip
Elton~Pinto$^{\text{*1}}$\enskip
Hakesh~Darapaneni$^{\dag\text{1}}$\enskip
Mitali Meratwal$^{\dag\text{1}}$\enskip
Shivam~Mittal$^{\dag\text{1}}$\enskip
Pranavi~Bajjuri$^{\dag\text{1}}$\enskip
Srinivas~Sridharan$^{\text{2}}$\enskip
Alexey~Tumanov$^{\text{1}}$
}

\affiliation{\vspace{1mm} $^{\text{1}}$Georgia Institute of Technology\enskip
$^{\text{2}}$NVIDIA Inc. \country{}}

\thanks{Equal technical contribution. Srihas Yarlagadda and Elton Pinto led the development on \sysname and \syssearchname respectively, Amey Agrawal was responsible for the overall system design. $^\dag$ Work done while at Georgia Institute of Technology.}

\maketitle
\renewcommand{\shortauthors}{Yarlagadda, Agrawal, Pinto et al.}
\input{1-intro}

\input{2-background}

\input{3-motivation}

\input{4-design}
\input{5-config-search}

\input{6-impl}

\input{7-eval}

\input{8-discussion}
\input{9-related}

\input{10-conc}

\input{11-ack}

\bibliographystyle{plain}
\bibliography{main}

\newpage

\input{11-appendix}

\end{document}

%% file: package.tex
\usepackage{tikz}
\usepackage{amsmath}
\usepackage{multirow}
\usepackage{flushend}
\usepackage{enumitem}
\usepackage{algorithm}
\usepackage{algpseudocodex}
\usepackage{subcaption}
\usepackage{hyperref}
\usepackage{xcolor,colortbl}
\algtext*{EndWhile} %
\algtext*{EndIf} %
\algtext*{EndFor} %

\usepackage{url}

\usepackage{breakurl}
\usepackage{hyperref}
\makeatletter
\renewcommand{\sectionautorefname}{\S\@gobble}
\renewcommand{\subsectionautorefname}{\S\@gobble}
\renewcommand{\subsubsectionautorefname}{\S\@gobble}
\makeatother
\hypersetup{
   colorlinks,
   linkcolor={blue!50!black},
   citecolor={red!50!black},
   urlcolor={blue!80!black}
}

\usepackage{comment}
\usepackage{xspace}
\usepackage{soul}
\usepackage{tikz}
\usepackage{tcolorbox}
\usepackage{array}
\usepackage{graphicx}
\usepackage{xparse}
\usepackage{booktabs}
\usepackage{lipsum}
\usepackage{ifthen}
\usepackage[font=small,labelfont=bf]{caption}
\usepackage[capitalize,nameinlink]{cleveref}

\setlist[itemize]{noitemsep, topsep=1pt}
\setlist[enumerate]{noitemsep, topsep=1pt}

%% file: macros.tex
\newcommand{\sysname}{\textsc{Maya}\xspace}
\newcommand{\syssearchname}{\textsc{\sysname-Search}\xspace}

\newcommand{\myx}{$\times$\xspace}
\newcommand{\csl}{custom specification languages\xspace}
\newcommand{\semg}{semantic gap\xspace}

\newcommand{\vheading}[1]{\vspace{0.02in}\noindent\textbf{#1.}}
\newcommand{\sref}[1]{\S\ref{#1}}

\newcommand{\evalpe}{5} %
\newcommand{\evaltc}{56} %
\newcommand{\evalcg}{2} %

\newcommand{\ctc}{74} %

\definecolor{dgreen}{rgb}{0.0, 0.5, 0.0}
\definecolor{codegreen}{rgb}{0,0.6,0}
\definecolor{codegray}{rgb}{0.5,0.5,0.5}
\definecolor{codepurple}{rgb}{0.58,0,0.82}
\definecolor{backcolour}{rgb}{0.95,0.95,0.92}
\definecolor{purple}{RGB}{128,0,128}
\definecolor{indigo}{RGB}{75,0,130}
\definecolor{royalblue}{RGB}{65,105,225}
\definecolor{navy}{RGB}{0,0,128}
\definecolor{codebrown}{rgb}{0.6,0.6,0}

\newcommand{\greencheck}{\textcolor{codegreen}{\checkmark}}
\newcommand{\redcross}{\textcolor{red}{$\times$}}

\newcommand{\greenuparrow}{\textcolor{codegreen}{$\uparrow$}}
\newcommand{\greendownarrow}{\textcolor{codegreen}{$\downarrow$}}
\newcommand{\reduparrow}{\textcolor{red}{$\uparrow$}}
\newcommand{\reddownarrow}{\textcolor{red}{$\downarrow$}}

\ifthenelse{\boolean{publicversion}}{
	\newcommand{\grumbler}[3]{}
        \newcommand{\rishabh}[1]{}
        \newcommand{\hakesh}[1]{}
        \newcommand{\mitali}[1]{}
        \newcommand{\shivam}[1]{}
        \newcommand{\elton}[1]{}
        \newcommand{\srihas}[1]{}
        \newcommand{\srinivas}[1]{}
        \newcommand{\alexey}[1]{}
        \newcommand{\amey}[1]{}
        \newcommand{\todo}[1]{\grumbler{XXX}{#1}{red}}
}
{
\newcommand{\grumbler}[3]{\xspace\textcolor{#3}{\bf #1: #2}}
\newcommand{\rishabh}[1]{\grumbler{Rishabh}{#1}{magenta}}
\newcommand{\hakesh}[1]{\grumbler{Hakesh}{#1}{magenta}}
\newcommand{\mitali}[1]{\grumbler{Mitali}{#1}{magenta}}
\newcommand{\elton}[1]{\grumbler{Elton}{#1}{magenta}}
\newcommand{\shivam}[1]{\grumbler{Shivam}{#1}{magenta}}
\newcommand{\srihas}[1]{\grumbler{Srihas}{#1}{violet}}
\newcommand{\srinivas}[1]{\grumbler{Srinivas}{#1}{teal}}
\newcommand{\alexey}[1]{\grumbler{Alexey}{#1}{cyan}}
\newcommand{\amey}[1]{\grumbler{Amey}{#1}{violet}}
\newcommand{\todo}[1]{\grumbler{XXX}{#1}{red}}
}

\newcommand{\amped}{AMPeD\xspace}
\newcommand{\calc}{Calculon\xspace}
\newcommand{\prot}{Proteus\xspace}

%% file: 0-abstract.tex
\begin{abstract}

Training large foundation models costs hundreds of millions of dollars, making deployment optimization critical. Current approaches require machine learning engineers to manually craft \textit{training recipes} through  error-prone trial-and-error on expensive compute clusters. To enable efficient exploration of training configurations, researchers have developed performance modeling systems. However, these systems force users to translate their workloads into \csl, introducing a fundamental \textit{\semg} between the actual workload and its representation. This gap creates an inherent tradeoff: systems must either support a narrow set of workloads to maintain usability, require complex specifications that limit practical adoption, or compromise prediction accuracy with simplified performance models.

We present \sysname, a performance modeling system that eliminates these tradeoffs through transparent device emulation. By operating at the narrow interface between training frameworks and accelerator devices, \sysname can capture complete workload behavior without requiring code modifications or translations. \sysname intercepts device API calls from unmodified training code to directly observe low-level operations, enabling accurate performance prediction while maintaining both ease of use and generality. Our evaluation shows \sysname achieves less than \evalpe{}\% prediction error across diverse models and optimization strategies, identifying configurations that reduce training costs by up to \evaltc{}\% compared to existing approaches.

\end{abstract}

%% file: 1-intro.tex
\section{Introduction}\label{sec:Introduction}

\input{figures-tex/banner}
\input{figures-tex/parallel_coords_strong_scaling_18b_h100}

Large foundation models like ChatGPT \cite{openai2023chatgpt} and Sora \cite{sora} have demonstrated human-level performance across various natural language and visual tasks. The development of these models critically depends on scaling both model sizes and training corpora \cite{kaplan2020scaling}. Consequently, the computational demands for training have reached staggering proportions. For instance, training the Llama-3 405B model required 54 days on 16,000 accelerators \cite{dubey2024llama}, a setup that would cost over \$250 million on the Microsoft Azure public cloud \cite{dubey2024llama}.

Training at this scale requires sophisticated system optimizations. Researchers have developed techniques spanning distributed training strategies (tensor, pipeline, expert parallelism) \cite{megatron,singh2023hybrid}, compute optimizations (kernel fusion, pipeline interleaving) \cite{megatron}, and memory optimizations (activation checkpointing, gradient accumulation) \cite{megatron,zero,korthikanti2023reducing}. Engineers meticulously craft \textit{training recipes} that combine these techniques to maximize hardware utilization. However, the vast array of techniques and their associated parameters create a combinatorial explosion in the configuration space.

Furthermore, these recipes need to be tailored for each deployment scenario. As depicted in \Cref{fig:motivation:strongscalingconfigs:configs}, even small changes in the deployment scenario can require significant alterations to the configuration. Applying a recipe optimized for one scenario to another can degrade efficiency by up to \ctc{}\%, as shown in \Cref{fig:motivation:strongscalingconfigs:costofmisconfig}.

These challenges prop up the need for efficient runtime modeling that can evaluate training strategies without requiring actual hardware deployment. Naive analytical models cannot capture the complex characteristics of these distributed training workloads, leading to inaccurate prediction. To address this challenge, several advanced runtime modeling systems \cite{duan2023proteus,bang2023vtrain,lu2023distsim,zhu2020daydream,isaev2023calculon,moolchandani2023amped,santhanam2021distir} have been proposed. However, these systems suffer from a fundamental limitation: they cannot operate directly on the user code, and require translating the workloads into \textit{\csl}.

This translation process introduces two fundamental critical challenges. Consider that a user wants to optimize a GPT-3 model run using one of the existing tools. To employ Proteus \cite{duan2023proteus}, an engineer must translate the original PyTorch workload into a ``Strategy Tree" format --- requiring hundreds of lines of specialized code \cite{proteusgithub} that explicitly encode parallelization patterns, communication topology, and memory optimizations. Any detail omitted or simplified during this manual translation leads to prediction errors (\Cref{fig:eval:fidelity:pointcloud}), resulting in up to \evaltc\% higher training costs (\Cref{fig:eval:fidelity:cost}). We term this loss of implementation detail during the translation process as the \textbf{\textit{semantic gap}}.

Second, system designers face an inherent \textbf{\textit{generality-usability tradeoff}}. Systems prioritizing generality like Proteus employ expressive but complex specifications, while systems optimizing for usability like Calculon \cite{isaev2023calculon} and AMPed \cite{moolchandani2023amped} offer simpler interfaces but only support specific frameworks like Megatron-LM, limiting their applicability.

We observe that while training systems are complex, they interact with accelerators through a narrow, well-defined interface of device APIs. Moreover, training workloads exhibit a fundamental property: control flow (executed on CPUs) rarely depends on specific numerical computation results (executed on accelerators). This decoupling is pervasive in modern training --- data-parallelism processes different data shards with identical control flow, and even techniques like gradient accumulation and mixed precision training maintain deterministic control patterns. While this excludes certain architectures with data-dependent control flow (e.g., some MoE implementations), these represent a small fraction of workloads found in the wild \cite{zhu2020daydream,geoffrey2021habitat,duan2023proteus,isaev2023calculon,lu2023distsim,flexflow,unger2022unity}.

We present \sysname, a transparent runtime modeling system that exploits these insights through \textit{transparent device emulation}. Rather than requiring workload translation, \sysname intercepts and emulates all accelerator API interactions from unmodified training code, creating the illusion of actual device execution while capturing complete workload behavior. This approach eliminates both the semantic gap and the generality-usability tradeoff. Our evaluation demonstrates prediction error below \evalpe{}\% across diverse configurations.

\noindent In summary, we make the following contributions:

\begin{itemize}[leftmargin=*,itemsep=3pt,topsep=3pt]
\item We identify intrinsic limitations of existing runtime modeling approaches for DL training workloads, specifically the semantic gap and generality-usability tradeoff arising from custom specification languages.
\item We propose \sysname, a transparent and flexible runtime modeling system that emulates workload execution on accelerated compute clusters.
\item We demonstrate that \sysname can predict the end-to-end runtime of workloads with < \evalpe{}\% error across a variety of models and training configurations.
\item Finally, we demonstrate the efficacy of our system for finding optimal training recipes, reducing training cost by up to \evaltc{}\% compared to existing systems.
\end{itemize}

%% file: figures-tex/banner.tex
\begin{figure}
    \small
    \centering
    \begin{subfigure}[b]{0.5\linewidth}
        \centering
        \includegraphics[width=\linewidth]{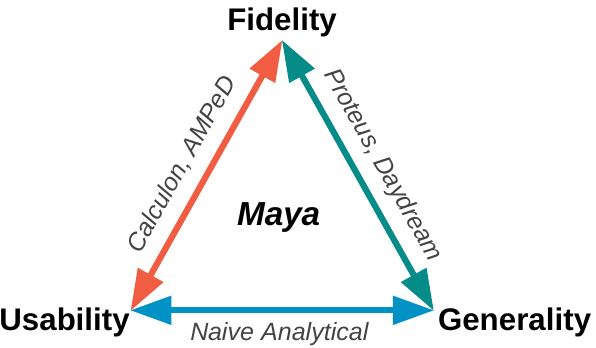}
        \caption{Trade-off Space.}
       \label{fig:banner:tradeoff}
    \end{subfigure}
    \begin{subfigure}[b]{0.4\linewidth}
        \centering
        \includegraphics[width=0.8\linewidth]{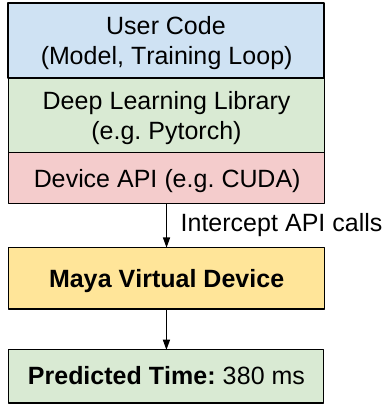}
        \caption{\sysname overview.}
       \label{fig:banner:overview}
    \end{subfigure}
    \caption{\small Existing deep learning training performance modeling systems struggle with a tradeoff between fidelity, usability, and generality. Naive analytical models lack fidelity, while the advanced modeling approaches make a tradeoff between usability and generality. \sysname breaks this tradeoff through a novel device emulation approach, achieving all three simultaneously.}
   \label{fig:banner}
\end{figure}

%% file: figures-tex/parallel_coords_strong_scaling_18b_h100.tex
\begin{figure*}[htbp]
    \centering
    
    \begin{subfigure}{0.69\textwidth}
        \centering
        \includegraphics[width=\textwidth]{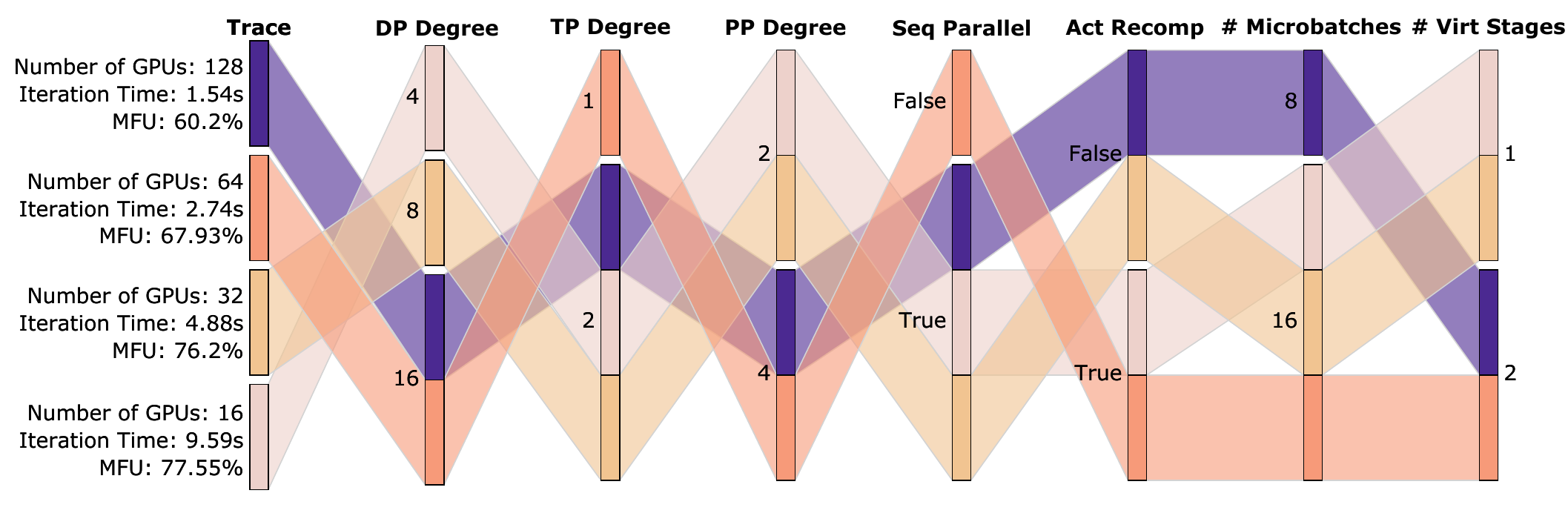}
        \vspace{0.5em}
        \caption{Configuration Shifts Across Cluster Sizes}
        \label{fig:motivation:strongscalingconfigs:configs}
    \end{subfigure}
    \hfill
    \begin{subfigure}{0.29\textwidth}
        \centering
        \includegraphics[width=\textwidth]{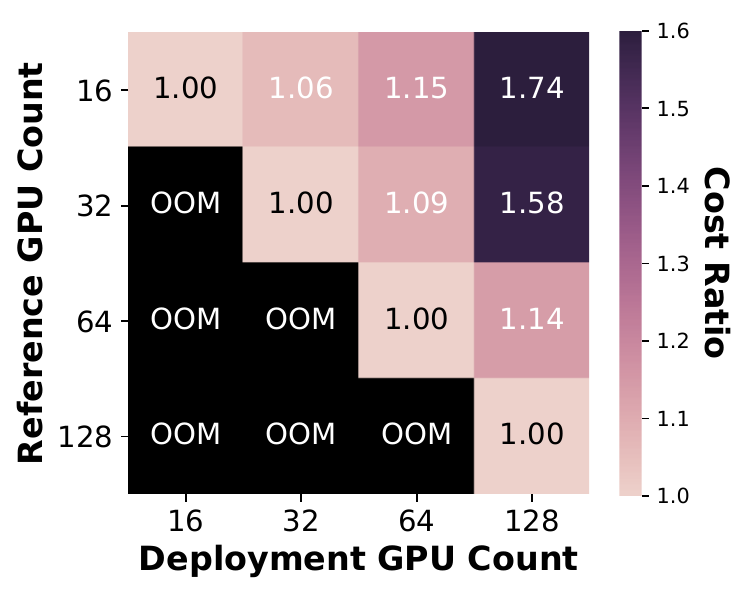}
        \vspace{0.1em}
        \caption{Cross-Deployment Inefficiency Matrix}
    \label{fig:motivation:strongscalingconfigs:costofmisconfig}
    \end{subfigure}
    \caption{Sensitivity of optimal training configurations to cluster size for GPT-3 18.4B on H100 GPUs. As GPU counts increase, configurations shift fundamentally --- from memory-efficient combinations of tensor and pipeline parallelism in smaller clusters to higher data-parallel degrees in larger clusters. The cross-deployment cost matrix highlights that deploying configurations tuned for one cluster size on another can lead to inefficiencies, increasing costs by up to \ctc{}\% due to suboptimal resource use. These results emphasize the necessity of scenario-specific configuration tuning, a key challenge \sysname addresses through precise performance modeling.}
\label{fig:motivation:strongscalingconfigs}
\end{figure*}

%% file: 2-background.tex
\input{tables/sys_comp}

\section{Background}
\label{sec:background}

Deep learning training (DLT) workloads have grown to unprecedented scales, with state-of-the-art models now containing billions of parameters. For instance, training the Llama-3 405B required an estimated 864 exaflop-days of compute \cite{dubey2024llama}. Training such large-scale deep learning models requires parallelizing computation across hundreds or thousands of accelerator devices. This parallelization employs techniques such as Data Parallelism (DP), which replicates the model across devices; Tensor Parallelism (TP), which partitions individual layers; and Pipeline Parallelism (PP), which splits the model into stages. However, parallelization alone is insufficient; achieving high efficiency requires carefully balancing compute, memory, and communication bottlenecks.

\vheading{Balancing Resource Utilization} To address these bottlenecks, researchers have proposed various techniques that trade off different resources (Table \ref{tab:knob_impacts}). For instance, tensor parallelism can reduce memory pressure by partitioning layers across devices, but increases communication overhead due to frequent all-reduce operations between partitioned layers. Pipeline parallelism \cite{gpipe,pipedream} introduces pipeline bubbles that reduce compute efficiency, but enables parallelization at a comparatively low communication cost and memory pressure. The Zero Redundancy Optimizer (ZeRO) \cite{zero} shards model parameters, gradients, and optimizer states across workers, reducing memory pressure at the cost of increased communication. Activation checkpointing \cite{shah2020memory} performs additional recomputation to reduce memory usage, dropping activations after the forward pass.

Efficient hardware utilization requires careful composition of these techniques based on the model architecture, training parameters, and available resources. Each technique introduces additional tunable parameters that affect this balance. For example, the interleaved 1F1B pipelining schedule \cite{pipedream} reduces pipeline bubbles by assigning multiple micro-batches to each pipeline stage, but requires careful tuning of micro-batch counts to balance communication overlap. Similarly, ZeRO offers different sharding stages that provide varying tradeoffs between memory and communication.

\vheading{Composing Training Recipes} The vast configuration space generated by these techniques, combined with their interdependencies, makes tuning DLT workloads challenging. \Cref{fig:motivation:strongscalingconfigs} illustrates optimal configurations for training the same model with varying numbers of accelerators. With limited resources (16 devices), a combination of tensor and pipeline parallelism alleviates memory pressure. When scaling to 128 devices, the reduced per-device memory requirement allows leveraging data parallelism instead of tensor parallelism, avoiding additional communication overhead. Our experiments show that using the optimal configuration for 16 devices results in a 1.74\myx higher cost when applied to 128 devices, compared to the optimal configuration. This performance sensitivity makes it impractical to rely solely on heuristics or previous experience for configuration selection.

\textit{\textbf{Takeaway:} DLT workloads employ a rich set of parallelization and optimization techniques with unique resource tradeoffs. These techniques must be carefully composed to maximize hardware utilization.}

\input{tables/parallelism_knob_impacts}

Given these challenges, we pose the following question: \textit{Can we transparently, accurately, and efficiently predict the performance of arbitrary DLT configurations without access to target hardware?} Answering this question is crucial for enabling rapid exploration of the configuration space to identify resource-efficient training recipes.

%% file: tables/sys_comp.tex
\begin{table*}[htbp!]
\small
    \centering
    \begin{tabular}{lcccccccc} 
    \toprule
     & \multicolumn{1}{c}{} & \multicolumn{4}{c}{\textbf{Domain Specific Simulators}} & \multicolumn{3}{c}{\textbf{Analytical Models}} \\ 
    \cmidrule(l{2pt}r{2pt}){3-6}\cmidrule(l{2pt}r{2pt}){7-9}
    & \textbf{\sysname} & Proteus & vTrain & DistSim & Daydream & Calculon & AMPed & DistIR \\
    \noalign{\hrule height 0.5pt}
    \rowcolor{gray!15}
         \multicolumn{9}{c}{\textit{System Properties}}\\\hline Deployment-Free Prediction  & \greencheck & \greencheck & \redcross & \redcross & \redcross & \greencheck & \greencheck & \greencheck \\
Transparent -- \textit{No Code Modifications} & \greencheck & \redcross & \redcross & \redcross & \redcross & \redcross & \redcross & \redcross \\
Workload Agnostic & \greencheck & \greencheck & \redcross & \greencheck & \greencheck & \redcross & \redcross & \greencheck \\
\noalign{\hrule height 0.5pt}
    \rowcolor{gray!15}
         \multicolumn{9}{c}{\textit{Modeling Domain}}\\\hline
    Data Parallel & \greencheck & \greencheck & \greencheck & \greencheck & \greencheck & \greencheck & \greencheck & \greencheck \\
    Tensor Parallel & \greencheck & \greencheck & \greencheck & \greencheck & \redcross & \greencheck & \greencheck & \greencheck \\
     Pipeline Parallel & \greencheck & \greencheck & \greencheck & \greencheck & \redcross & \greencheck & \greencheck & \greencheck \\
    Sequence Parallel & \greencheck & \redcross & \redcross & \redcross & \redcross & \greencheck & \redcross & \redcross \\
    Pipeline Interleaving & \greencheck & \greencheck & \redcross & \redcross & \redcross & \greencheck & \redcross & \greencheck \\
    Distributed Optimizer & \greencheck & \greencheck & \redcross & \redcross & \redcross & \greencheck & \redcross & \greencheck \\
    Activation Recomputation & \greencheck & \greencheck & \redcross & \redcross & \redcross & \greencheck & \redcross & \redcross \\
    Gradient Accumulation & \greencheck & \redcross & \redcross & \redcross & \redcross & \greencheck & \redcross & \greencheck \\

    \bottomrule
    \end{tabular}
    \caption{
    Comparison of \sysname with existing performance modeling approaches. \sysname uniquely combines deployment-free prediction, transparency, and workload agnosticism, supporting a broad range of parallelism and optimization strategies across training configurations. Competing systems are limited in either coverage and often require code modifications.}
    \label{tab:features_systems}
\end{table*}

%% file: tables/parallelism_knob_impacts.tex
\begin{table}[htbp]
    \centering
    \begin{tabular}{@{\hspace{1px}}l@{\hspace{0px}}c@{\hspace{4px}}c@{\hspace{4px}}c@{\hspace{1px}}} 
    \toprule
    \textbf{Resource Load} & Compute & Memory & Network \\
    \midrule
    Data Parallel & \reddownarrow & \greendownarrow & \reduparrow \\
    Tensor Parallel & \reddownarrow & \greendownarrow & \reduparrow \\
    Pipeline Parallel & \reddownarrow & \greendownarrow & \reduparrow  \\
    Sequence Parallel & \reddownarrow & \greendownarrow & \reduparrow \\
    Pipeline Interleaving & \greenuparrow & \greendownarrow & \reduparrow \\
    Distributed Optimizer & \textbf{--} & \greendownarrow & \reduparrow \\
    Activation Recomputation & \reddownarrow & \greendownarrow & \textbf{--} \\
    Gradient Accumulation & \reddownarrow & \greendownarrow & \greendownarrow \\
    \bottomrule
    \end{tabular}
    \caption{Effect of configuration knobs on compute utilization, memory load, and network load in a fixed global batch size setting. This table highlights the trade-offs associated with each knob: while some configurations increase compute utilization, others may reduce memory or network load, illustrating the balancing act required to optimize  large-scale training jobs. \vspace{-1.5em}
    }
    \label{tab:knob_impacts}
\end{table}

%% file: 3-motivation.tex
\section{Challenges \& Key Idea}
\label{sec:motivation}

As modern deep learning training (DLT) workloads scale to unprecedented levels, it has become increasingly important to optimize their resource allocation, cost, and environmental impact. Runtime performance prediction systems vastly aid such optimization efforts; however, the complexity of modern DLT workloads (involving clusters with hundreds-thousands of GPUs) and the use of domain-specific optimizations makes developing such systems quite challenging.

At a high level, state-of-the-art runtime modeling systems~\cite{duan2023proteus,isaev2023calculon,bang2023vtrain,lu2023distsim,zhu2020daydream,geoffrey2021habitat,moolchandani2023amped,santhanam2021distir} follow a four-phase approach to predict DLT workload performance (\Cref{fig:simanatomy}):

\begin{enumerate}
\item \textbf{Workload Specification:} Translate the DLT job into a framework-specific representation, capturing computational flow of the workload.
\item \textbf{Kernel Decomposition:} Decompose the workload into an execution graph containing the kernels.
\item \textbf{Kernel Runtime Prediction:} Estimate execution times for the individual kernels using analytical models or historical profiling data.
\item \textbf{Distributed Execution Simulation:} Model the end-to-end execution, accounting for inter-device communication and synchronization.
\end{enumerate}

There is, however, an inherent flaw in this approach --- it lacks transparency. Users have to explicitly encode their workload in a custom out-of-band specification language, which then has to be decomposed into a kernel execution graph. This undermines the efficacy of such systems due to two interrelated issues: (1) the custom specification language may be insufficiently expressive or difficult to use (\sref{sub:usability-tradeoff}), and (2) the workload specification may not accurately represent the true workload (\sref{sub:semantic-gap}). Further, even if it were possible to devise a language that is both expressive and easy-to-use, lack of transparency makes such systems \textit{fragile} and \textit{inflexible}. Users are required to revisit existing specifications and devise new ones as DLT workloads evolve. Therefore, there is a clear need for a transparent, user-friendly, and accurate runtime modeling system.

\subsection{The Generality-Usability Tradeoff}
\label{sub:usability-tradeoff}

Existing systems attempt to maintain fidelity while navigating this semantic gap through two primary approaches. On one end of the spectrum, systems like DistIR~\cite{santhanam2021distir} and Proteus~\cite{duan2023proteus} opt for highly expressive but complex representation formats. While these can capture intricate details of the workload, they require users to translate their jobs into hundreds of lines of specialized code~\cite{distirgithub, proteusgithub}. This imposes a substantial burden on users and, consequently, introduces opportunities for translation errors that can compromise prediction accuracy.

On the other end, systems like VTrain~\cite{bang2023vtrain} and Calculon~\cite{isaev2023calculon} prioritize usability by providing simpler interfaces where users only need to specify configuration parameters. However, this simplicity comes at the cost of generality. These systems are tightly coupled to specific workload implementations, such as Megatron-LM~\cite{megatron}, limiting their applicability to a narrow range of use cases.

As a result, there is a tension between usability and generality in current approaches. Systems that strive for broad applicability often sacrifice ease of use, while those focusing on user-friendliness sacrifice generality.

\subsection{Semantic Gap in Workload Representation}
\label{sub:semantic-gap}

Existing runtime prediction frameworks require users to manually encode their workloads either using out-of-band \csl or through static configuration knobs. This approach makes it more likely for a \textit{semantic gap} to manifest between the actual workload and its abstract representation, leading to inaccurate predictions. For instance, users may inadvertently omit complex hardware-specific optimizations or subtle system interactions. Additionally, many DLT workloads exhibit complex runtime behaviors that are non-trivial to represent. This is an inherent ``garbage in, garbage out'' problem --- inaccurate workload specifications lead to inaccurate predictions.

\subsection{Illustrative Example}

Consider the scenario presented in \cref{fig:motiv:example}. AMPed restricts users to a fixed set of operators with carefully curated analytical models. While these analytical predictions can be composed to produce final runtime estimates, the rigid modeling language introduces significant approximation errors. In contrast, Proteus provides an expressive intermediate representation (IR) that allows users to encode diverse parallelism schemes through strategy trees. However, this flexibility comes at a cost --- users may inadvertently model features incorrectly, and verifying that the translated model accurately represents the original program is a challenging and error-prone process.

These limitations are apparent when attempting to evaluate a new framework optimization. A pertinent example is DualPipe \cite{dualpiperepo}, a pipeline parallelism schedule utilized in the training of DeepSeek-R1 \cite{deepseek}. This schedule differs from the usual interleaved 1F1B schedule proposed by Megatron-LM in that it increases overlapping by running pairs of micro-batches bidirectionally. A static analysis approach would require relevant calculations for the forward and backward passes --- specifically those accounting for the pipeline bubble --- to be rewritten to reflect the increased overlap. On the other hand, expressing this schedule with Proteus would require a custom graph transformation pass on the strategy tree to introduce additional compute/communication nodes, which is a manual and cumbersome process.
\input{figures-tex/example}

\subsection{Solution: Transparent Device Emulation}

To address the absence of transparent abstractions, we propose a novel solution that leverages the unique characteristics of DLT workloads to achieve generality and ease of use without sacrificing fidelity: \textit{transparent device emulation}. Instead of having the user specify their workload in an obtuse specification language, we intercept and emulate all application interactions with the accelerator. The system then simulates cluster behavior from these intercepted traces, yielding a performance prediction.

An emulation-driven approach is viable for two key reasons. First, despite the complexity of DLT workloads, these systems interact with accelerators through narrow-waist, well-defined accelerator APIs. This allows \sysname to mimic the functionality of device management API calls such as \texttt{cudaMalloc}, \texttt{cudaSetDevice}, and \texttt{cublasSetMatrix}, creating the illusion that the user application is running on the actual device. Second, DLT applications exhibit a critical decoupling between control flow (executed on the CPU) and computation (executed on the accelerator device), with the former rarely depending on the actual computation results. This separation allows us to emulate device execution without affecting the application's control flow. We simply save metadata for each compute operation but skip their actual execution, enabling rapid and resource-efficient tracing.

There are several benefits to this approach. First, by transparently intercepting accelerator interactions, \sysname is able to \textbf{accurately capture} the entire workload behavior \textbf{without requiring any changes to application code}. This on its own addresses the generality-usability trade-off and semantic gap. Users can now model DLT workloads without an intermediate workload encoding step, significantly lowering the barrier to adoption. The transparency of \sysname makes it highly adaptable and resilient to the ever-evolving landscape of DLT workloads.

Second, the detailed workload trace from emulation can be used to produce high-fidelity predictions (\sref{sec:eval:fidelity}). Downstream processing and simulation can accurately represent low-level behavior, making it easy to identify bottlenecks and generate reliable predictions. Further, each component of the \sysname stack --- emulation, trace processing, runtime estimation, and simulation --- \textbf{is pluggable and can be tuned separately}, enabling improved overall accuracy of runtime predictions and better flexibility, all while remaining transparent.

Finally, \sysname is fundamentally \textit{not coupled} to a specific framework or model, allowing it to seamlessly integrate with existing workflows. It can also be used to build more sophisticated systems such as configuration search (\sref{sec:eval:opti}).

To summarize, \sysname is a transparent, user-friendly, and accurate performance modeling system designed to predict DLT workload behavior without requiring access to accelerator hardware. The key insight behind \sysname's design is that by operating at the narrow interface between training frameworks and accelerator devices, we can eliminate the fundamental trade-offs between modeling accuracy, ease of use, and generality.

%% file: figures-tex/example.tex
\begin{figure}[htbp]
    \centering
    \includegraphics[width=\linewidth]{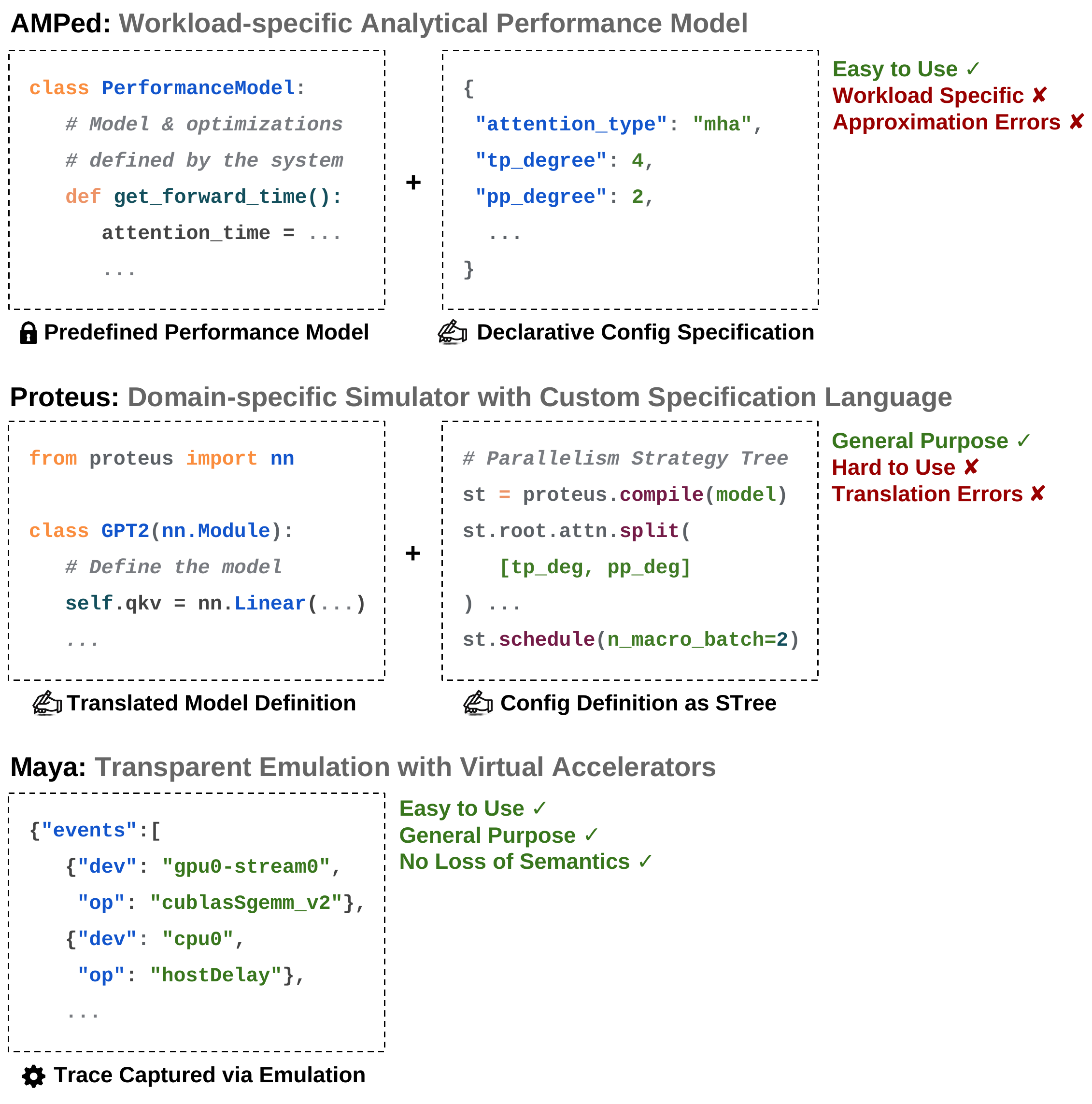}
    \caption{The user workflow across three systems. With AMPed (top), the user provides a declarative configuration specifying high-level parameters which are then fed into a predefined analytical model in the system. If a new model architecture or performance optimization is introduced, the system is rendered unusable. In Proteus (middle), the user must manually translate their entire model into a custom format and write a separate ``Strategy Tree'' to explicitly define the parallelization strategy. With \sysname (bottom), the user runs their original, unmodified training script, and the system automatically captures a low-level execution trace through transparent emulation, requiring no user intervention. \vspace{-1em}}
    \label{fig:motiv:example}
\end{figure}

%% file: 4-design.tex
\input{figures-tex/simulator-anatomy}

\input{figures-tex/hld}

\vspace{-0.5em}

\section{\sysname : System Design}\label{sec:design}

\input{figures-tex/trace_flow}

The foundation of \sysname is a transparent device emulator that functions as the interface between unmodified training workloads and the modeling pipeline. This component interposes on the accelerator device APIs, virtualizing device interactions while maintaining execution fidelity. The emulator captures a precise trace of device operations --- compute kernels, memory operations, and synchronization events --- completely on the CPU. This yields detailed execution traces while remaining fully transparent to the training application, which executes as if on a real accelerator.

These raw execution traces then flow through a trace collation and analysis pipeline that reconstructs the distributed execution pattern. The collator combines traces from multiple workers, resolving dependencies across both space (between workers) and time (within execution streams) by identifying collective communication operations, which are crucial for modeling distributed training workloads. The performance estimation phase then augments this execution trace with runtime predictions. Since the emulator captures operation metadata but does not execute compute kernels, \sysname employs a combination of machine learning and analytical models to predict operation runtimes.

The final phase uses event-driven simulation to model end-to-end execution. The simulator processes the annotated trace according to a specified hardware configuration, modeling complex execution dependencies within and across workers in a distributed training workload. This captures critical performance characteristics like pipeline bubbles and compute-communication overlap that emerge from the interaction between device operations. The output is a comprehensive simulation report that encompasses metrics such as batch execution time, communication time \& memory usage.

\sysname's architecture enables it to capture the full complexity of modern DLT optimizations while providing high-fidelity performance predictions. By operating on unmodified user code and eliminating the requirement for accelerator hardware during prediction, \sysname offers a unique combination of transparency and efficiency. In the rest of this section, we provide the design details of each component in \sysname (Figure \ref{fig:hld}).

\subsection{Transparent Accelerator Emulation}
We make a key observation on the nature of deep learning training (DLT) workloads: the CPU-side control flow of the application is fundamentally decoupled from the computation executed on accelerator devices. Since there is minimal feedback to the control flow from the results of device operations, it is possible to emulate device behavior without materializing output values. \sysname's emulator exploits this characteristic --- turning compute operations into no-ops while carefully managing device state and dependencies.

To achieve transparency, the emulator \textit{intercepts} calls to device APIs without requiring modifications to the training application. Most device operations, particularly compute kernels, are transformed into no-ops that record metadata about the operation and return immediately. However, \sysname must still precisely track  device state to ensure correct execution. For instance, when applications query device state through APIs like \texttt{cudaMemGetInfo}, \sysname returns carefully constructed responses that mimic device behavior, allowing frameworks like PyTorch to make memory management decisions as they would on real hardware.

The design of this architecture addresses three key challenges in maintaining execution fidelity despite no-op execution: (1) maintaining the semantic meaning of API sequences, (2) tracking both physical and virtual resources, and (3) handling distributed dependencies. These challenges guide \sysname's semantically-aware emulation layer.

\vheading{Context-aware Operation Modeling} Several device operations gain meaning only when considered within the context of a broader sequence of API calls, requiring careful state tracking. In CUDA, for instance, \texttt{cudaStreamWaitEvent()} synchronizes ops across compute streams based on events recorded by \texttt{cudaEventRecord()}. We maintain a map of device state to model these dependencies correctly, even though the underlying operations don't execute.

A similar treatment is required for operations involving opaque libraries like cuBLAS and cuDNN, where configurations are built incrementally. For instance, a cuBLAS matrix multiplication involves a sequence of setup calls (\texttt{cublasSetMatrix()}, \texttt{cublasSetStream()}) before the actual computation (\texttt{cublasGemmEx()}). \sysname tracks these stateful API sequences to construct the complete operation metadata, essential for modeling performance-critical operations like matrix multiplications and convolutions.

\vheading{Resource Tracking} \sysname maintains a dynamic mapping of both physical and virtual resources during emulation. For memory management, \sysname tracks allocations and deallocations, allowing it to simulate real hardware constraints and detect errors such as out-of-memory (OOM) conditions and invalid memory accesses. In unified memory configurations, \sysname tracks tensor locations across host and device spaces and resolves ambiguity in API calls like \texttt{cudaMemcpyAsync} to accurately model workload behavior. 

In addition to physical resources, \sysname creates and manages virtual resources and handles that are returned to the application; examples of this include device handles, CUDA streams, and CUDA events. Any misconfiguration or user error --- such as using an invalid stream or an uninitialized descriptor --- is identified and flagged by \sysname using each handle's state. Through detailed accounting of both physical and virtual resources, \sysname provides a realistic foundation to emulate device behavior and potential failure scenarios; this is a key benefit unique to emulated tracing.

\vheading{Inter-Device Dependencies} In distributed deep learning, collective communication operations are used to synchronize data across devices. To emulate them accurately, \sysname captures the full lifecycle of these collectives. Each worker initializes a communicator using an API like \texttt{ncclCommInitRank}, which assigns ranks and defines the communication topology for operations like \texttt{ncclAllReduce}. This setup enables \sysname to accurately track data dependencies and the role of each worker device within the collective operation.

Once initialized, these communicators facilitate data transfers that often run on dedicated streams to achieve compute-communication overlap. For example, in \texttt{ncclAllReduce}, each device concurrently performs compute tasks on one stream while executing collective communication on another. \sysname models this behavior by simultaneously tracking communication and compute streams, enabling it to accurately capture blocking dependencies and the resulting overlap between computation and data transfer. For more complex parallelism patterns, such as 3D parallelism, \sysname tracks multiple communicators operating across different workload dimensions --- assigning unique identifiers to each one and logging associated events in the trace. Just like compute operations, there is no need to actually share data between worker CPU processes since the control flow does not depend on the result of the collective; this obviates the need for IPC and synchronization in the emulator.

This approach works particularly well for DLT workloads due to their predictable, repetitive nature. The training loop typically executes the same sequence of operations repeatedly in each iteration, with control-flow decisions rarely depending on specific numerical results from device computation. By exploiting this characteristic while carefully maintaining device state, \sysname can accurately model workloads without executing device operations. The emulator produces detailed traces that capture the full complexity of device interactions while remaining lightweight and efficient.

\subsection{Trace Collection and Analysis}

\vheading{Worker Trace Generation} \sysname captures detailed execution traces for each worker in the distributed training job. Rather than just logging API calls, we maintain rich context about each operation. For compute kernels, we record essential metadata including input/output tensor shapes, data types, and memory layouts --- information critical for runtime prediction. For instance, in a transformer layer's attention computation, we track matrix dimensions and sparsity patterns that significantly impact performance.

Each trace entry also includes precise timing of CPU-side operations between kernel launches, capturing essential host overhead and dispatch latency. We achieve this by measuring wall-clock deltas between API calls during emulation. This is particularly important when operating with state-of-the-art devices like NVIDIA H100s, where dispatch overhead can be significant, especially in workloads with many small kernels.

\vheading{Trace Collation} A key challenge in analyzing distributed training workloads is reconstructing the global execution pattern from individual worker traces. While individual workers are aware of the number of participants in a collective operation, they do not have visibility into \textit{which} workers are involved or the topology of the communication graph. The collator identifies collective operations (like \texttt{ncclAllReduce}) and matches them across workers using communicator IDs and sequence numbers. This allows us to reconstruct and model the full communication pattern faithfully.

\label{sec:deduplication}
\vheading{Optimization: Worker Deduplication} An insight that enables \sysname to efficiently scale to large distributed workloads is that many workers in DLT perform identical work. For instance, in data parallel training, each worker executes the same computation on different data shards. We exploit this pattern through dynamic worker deduplication. During the first training iteration, we profile all workers to establish operation patterns. We compute rolling hashes of operation sequences, allowing us to identify workers performing redundant computation. Upon detecting duplicates, we terminate redundant workers and continue profiling only the unique ranks. The trace collator later reconstructs the full execution pattern using these profiled ranks.

This optimization is particularly effective for large-scale training jobs. For example, in a 64-GPU job with 8-way TP and 8-way DP, we only need to profile a single worker since tensor and data parallel workers exhibit identical behavior.

\subsection{End-to-end Simulator}

The emulator trace contains metadata for each operation but lacks execution times, since operations are emulated and not dispatched to actual hardware. To produce an end-to-end performance estimate from this trace, the simulator pipeline i) predicts per-operation runtimes from metadata recorded during emulation, and ii) conducts a discrete-event simulation of cluster behavior.

\vheading{Kernel Runtime Estimation} \sysname’s kernel runtime estimators are pluggable components that estimate latency and bandwidth for \textit{individual} compute and collective operations. Users can provide any runtime estimator of their choosing for any kernel type (eg. Habitat~\cite{geoffrey2021habitat}, GPU-Mangrove~\cite{braun2020mangrove}, static-analysis based approaches~\cite{alavani2018predstatic}).

By default, \sysname uses random forest regressors trained on profiling data from kernel microbenchmarks, similar to prior approaches \cite{zhu2020daydream, vidur}. For collective operations, the reference estimators leverage profiling data of intra-host and inter-host link characteristics, considering varying data sizes and the topology of participating devices. Please refer to Appendix \ref{appendix:predictions} for more details.

To facilitate easy onboarding of new operations, \sysname offers a transparent \textit{profiling} mode that dispatches operations on real hardware (rather than emulating them), logging each operation’s arguments and observed runtime. This enables us to progressively build and integrate prediction models from production workloads.

\vheading{Resource Model} The simulator models both host and accelerator resources. Each host machine is represented by a dispatch queue that processes operations and manages device interactions. Each accelerator manages multiple execution streams and models concurrent operation processing. Synchronization operations like \texttt{cudaDeviceSynchronize} and \texttt{cudaStreamWaitEvent} are modeled using blocking waits in the corresponding streams. Host-side computation and launch overheads are also modeled as blocking operations in the dispatch queue using measurements from the emulation phase. \sysname shares its overall discrete-event simulation approach with prior work \cite{zhu2020daydream, santhanam2021distir} - however, we can capture fine-grained dependencies at the CUDA API granularity owing to the detailed traces collected via emulation. For more details, please refer to Algorithm \ref{alg:simulator}, Appendix \ref{appendix:simulator_core}.

\vheading{Network Model} Network operations are implemented using a global waitmap where participating devices register themselves, blocking their respective streams until all workers join. This waitmap can capture pipeline bubbles and effectively model compute-comms overlap --- data dependencies manifest as stalls on the corresponding accelerator stream, while any concurrent compute streams can proceed to the next event (\autoref{fig:traceflow}). This behavior is described in more detail in Algorithm \ref{alg:waitmaps}, Appendix \ref{appendix:simulator_core}.

After all participants join, the on-the-wire duration of each collective operation is a black-box prediction from the corresponding kernel runtime estimator. This abstracts any topology-dependent runtime effects into a single discrete event that is separate from other dataflow dependencies in the simulator, allowing network operations to be modeled in isolation. This allows users to choose between profiled collective data from their target cluster (nccl-tests), or network simulators like ASTRA-sim \cite{won2023astra}.

By reproducing the behavior of accelerator primitives, the simulator provides an accurate representation of cluster behavior. Operation-level modeling ensures we can capture fine-grained behavior while remaining general; new computational optimizations can be captured without additional effort. While our implementation targets CUDA devices, the design generalizes to other accelerators.

%% file: figures-tex/simulator-anatomy.tex
\begin{figure*}
    \small
    \centering\includegraphics[width=0.7\linewidth]{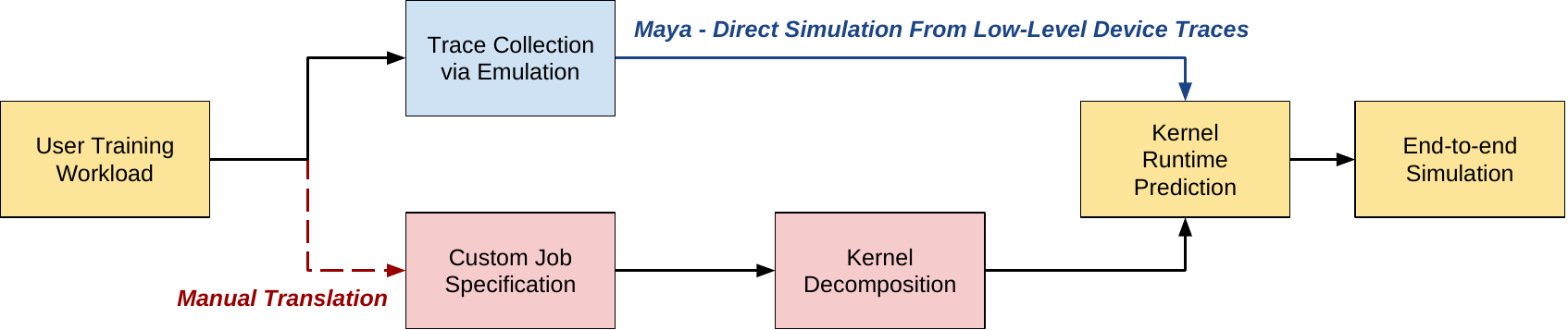}
    \caption{Comparison of modeling approaches: Traditional systems require explicit workload specification and several complex heuristics steps to obtain a kernel level execution graph apt for simulation. On the other hand, \sysname directly captures the computation graph at a lower-level through transparent device emulation.}
    \label{fig:simanatomy}
\end{figure*}

%% file: figures-tex/hld.tex
\begin{figure}[htbp]
    \centering
    \includegraphics[width=0.55\linewidth]{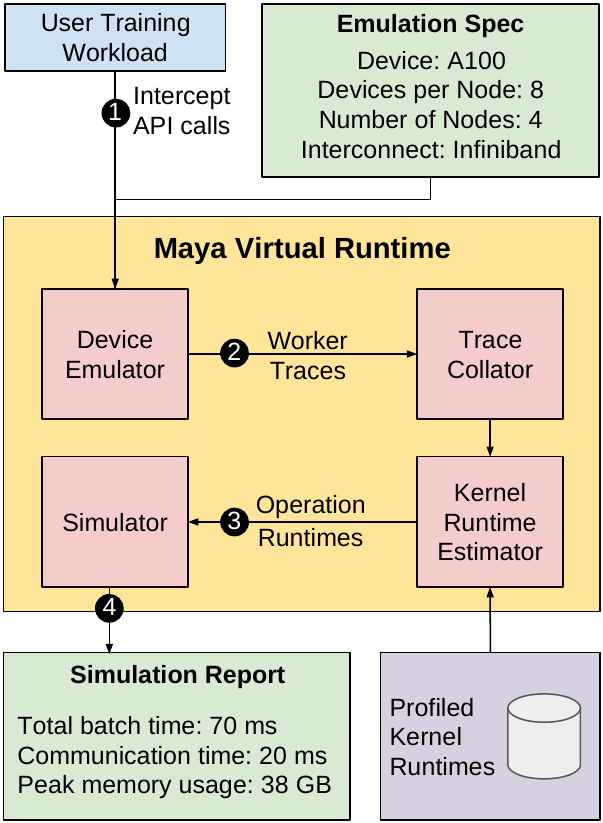}
    \caption{\sysname architecture: (1) Unmodified training code is executed through a virtual runtime that emulates the device drivers given emulation specs, (2) Worker traces capturing device API calls are merged into a unified trace, (3) Kernels in the unified trace are annotated with predicted runtimes using pre-trained estimators, (4) An event-driven simulator processes the annotated traces using cluster specifications to produce performance predictions.}
    \label{fig:hld}
\end{figure}

%% file: figures-tex/trace_flow.tex
\begin{figure*}
    \centering
    \includegraphics[width=\linewidth]{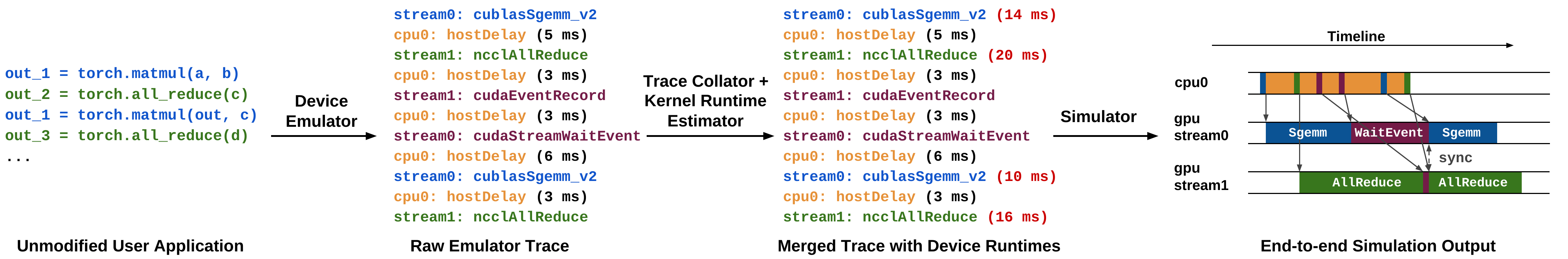}
    \caption{\small \sysname's trace processing pipeline: Starting with unmodified user code, the device emulator captures raw traces containing API calls, kernel launches, and synchronization events across multiple GPU streams. The trace collator merges these per-GPU traces and resolves collective operations, creating a unified job-level trace. The kernel runtime estimator then annotates compute operations with predicted durations. Finally, the event-driven simulator processes this trace to model the complex interactions between compute operations, synchronization events, and communication collectives across streams and devices, producing an accurate timeline of execution.}
    \label{fig:traceflow}
\end{figure*}

%% file: 5-config-search.tex
\vspace{-1em}
\section{Workload Tuning with \syssearchname}

\sysname's transparent emulation enables efficient exploration of the vast configuration space of DLT configs. While prior approaches require explicit modeling of each optimization technique, our emulation-based design naturally captures the impact of any configuration change through its low-level tracing. We leverage this capability to build an automated configuration search system that can rapidly evaluate different training recipes without requiring GPU resources.

The key insight is that by operating at the accelerator API level, \sysname can accurately predict the performance impact of configuration changes without needing to understand their semantic meaning. This allows us to treat configuration search as a black-box optimization problem, evaluating arbitrary combinations of parallelization strategies and system optimizations through lightweight emulation. The system takes as input a configuration space specification (defining the parallelization strategies and optimization knobs to explore), a resource specification (describing the target GPU cluster), and the training script. It then orchestrates concurrent trials that use \sysname to evaluate different configurations, continuously refining the search based on predicted performance using standard hyperparameter optimization techniques such as Bayesian optimization \cite{baysianoptimization} or Covariance Matrix Adaptation Evolution Strategy (CMA-ES) \cite{cma}.

\vspace{-1em}

\subsection{Concurrent Trial Scheduling} 

While \sysname's emulation engine provides a cheap way to evaluate different configurations, the search process can still be prohibitively slow if done sequentially. This necessitates careful resource management to enable the concurrent evaluation of multiple configurations. We solve this problem by developing a CPU scheduler that distributes the emulation of concurrent trials across CPU cores.

Since \sysname relies on wall-clock measurements for host-side overheads, concurrent trials would be affected by interference if they contend for CPU resources. To avoid this, we i) pin individual worker processes to CPU cores, and ii) run each emulated worker to completion before switching.

Second, we tackle memory pressure through careful process management. Each emulated GPU rank initially requires a complete copy of the user libraries (for instance, PyTorch runtime stack), which can quickly exhaust system memory when running several concurrent trials. We address this using Python's \textit{forkserver} mechanism to maintain a single copy of user libraries across workers, reducing memory footprint.

\input{figures-tex/point_cloud}

\subsection{Fidelity-Preserving Trial Pruning}

While concurrent execution improves throughput, we can further accelerate the search by intelligently pruning or skipping configurations that are guaranteed to perform worse than already evaluated ones. The key challenge is determining when such skips preserve prediction fidelity without missing potentially optimal configurations.

We develop a domain-aware trial scheduler that leverages known relationships between training configurations. For instance, if a configuration with activation recomputation enabled leads to out-of-memory (OOM) errors, we can safely skip evaluating the same configuration with recomputation disabled, as it will necessarily consume more memory and thus OOM. These relationships form a partial ordering over configurations based on their resource consumption.

The trial scheduler maintains a history of evaluated configurations and employs a set of conservative tactics to identify configurations that are dominated by previously seen ones. Pruning using this configuration history is \textit{fidelity-preserving}, meaning that no potentially optimal configuration is skipped while still achieving significant reductions in search time.

%% file: figures-tex/point_cloud.tex
\begin{figure*}[t!]
    \centering
    \includegraphics[width=0.8\textwidth]{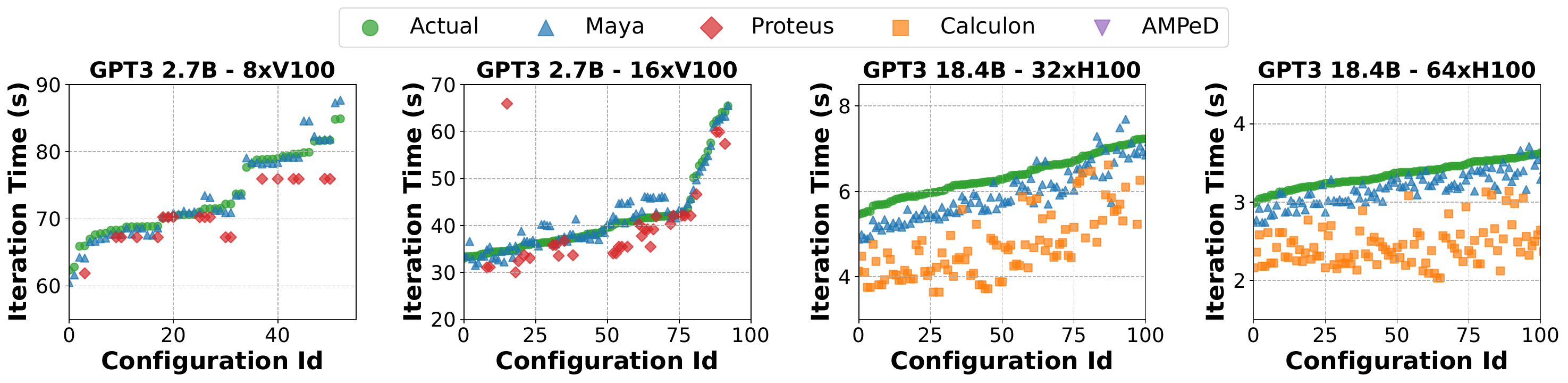}
    \caption{Runtime prediction accuracy comparison across different scales and hardware. We evaluate GPT3-2.7B (left) and GPT3-18.4B (right) models on V100 and H100 clusters. For each hardware setup, we plot the predicted vs actual per-iteration runtime for the top 100 valid configurations ranked by measured performance. \sysname consistently achieves high prediction fidelity across model sizes and hardware setups compared to existing approaches, with most predictions falling within 5\% of measured values. \todo{regen the plot with maya name}}
    \label{fig:eval:fidelity:pointcloud}
\end{figure*}

%% file: 6-impl.tex
\section{Implementation}

\sysname's CUDA emulator is implemented as a shared library ($\sim$2,500 lines in C++) that intercepts GPU-related API calls through dynamic linking. We use \texttt{LD\_PRELOAD} to inject our library at runtime, replacing symbols for the CUDA runtime API, driver API, and related libraries (cuBLAS, cuDNN, NCCL) with our implementations. This is similar to prior work on GPU virtualization \cite{singularity}. The event-driven simulator is implemented in Python ($\sim$3,000 lines) using a priority queue to process operation timings. The simulator includes specialized handlers for different operation types (compute, memory transfers, synchronization) and a topology-aware network model for accurate collective operation simulation. The configuration search system extends Ray Tune \cite{ray-tune} with domain-specific optimizations. The system exposes a simple Python API to integrate \syssearchname in less than 15 lines of code changes and  abstracts the complexity of emulation and trial management.

%% file: 7-eval.tex
\section{Evaluation}
\label{sec-eval}

In this section, we present a comprehensive evaluation of \sysname to demonstrate its effectiveness in predicting LLM training performance and optimizing deployment configurations. Our evaluation aims to answer these key questions:

1. How accurate is \sysname in predicting end-to-end runtime of training workloads across models of various sizes, different configurations and deployment environments?  (\sref{sec:eval:fidelity})

2. Can \sysname effectively optimize DLT workload deployment while navigating large configuration spaces? (\sref{sec:eval:opti})

3. How can \sysname scale to large clusters?(\sref{sec:eval:hyperscale})

To answer these questions, we conduct a series of experiments, where we compare \sysname's predictions against real-world measurements and existing state-of-the-art runtime modeling systems. Finally, we also present ablation studies to evaluate the effectiveness and scalability of different components of \sysname (\sref{sec:eval:ablation}).

\input{figures-tex/relative_cost}

\subsection{Experimental Setup}

\vheading{Baselines} We compare \sysname against a variety of state-of-the-art runtime modeling systems. We consider two analytical modeling frameworks -- Calculon \cite{isaev2023calculon} and AMPed \cite{moolchandani2023amped}, and one domain specific simulator, Proteus \cite{duan2023proteus}. We omit vTrain \cite{bang2023vtrain}, DistSim \cite{lu2023distsim} and Daydream \cite{zhu2020daydream} from comparison due to unavailability of their source code. While DistIR \cite{santhanam2021distir} is available publicly, it only supports modeling the training performance for simple MLP workloads.

\vheading{Models}  In order to facilitate a direct comparison, we conduct our experiments on the GPT-3 \cite{brown2020language} family of models -- the only workload natively supported by our baselines AMPed and Calculon. We use Megatron-LM \cite{megatron} GPT-3 \cite{brown2020language} 2.7B, 18.4B and 145.6B models in our experiments, with fixed global batch sizes of 256, 512 and 12k respectively (unless otherwise mentioned). The training scripts use HuggingFace Accelerate \cite{hfaccelerate}, Pytorch 2.1.0 \cite{ansel2024pytorch} and bfloat16 mixed precision. We also verify \sysname on a host of models using FSDP and \texttt{torch.compile} (Table \ref{tab:framework_generality}).

\vheading{Hardware} We evaluate the performance of \sysname in three different scenarios -- a 64 GPU NVIDIA H100 DGX \cite{h100} cluster, a 16 GPU V100 \cite{v100} DGX cluster, and a node containing 8 A40 GPUs. Each DGX-H100 server has 8 NVIDIA H100 GPUs with 80GB of High Bandwidth Memory (HBM). GPUs within a server are connected with NVLINK4.0 providing 900GBps bidirectional bandwidth. GPUs across servers are connected via Ethernet with RoCE offering 400Gbps per GPU pair. 

The V100 DGX servers are equipped with 8 GPUs with 40GB HBM memory capacity. Intra-node NVLINK connectivity is in an asymmetric cubemesh topology \cite{v100} with 300GBps links. These machines are connected using a 100GBps Infiniband \cite{infiniband} link. The A40 node uses pairwise NVLINK 4.0 between GPUs. Finally, we run the \sysname prediction pipeline on a CPU-only node (AMD 7513 EPYC, 128 cores, 504GB RAM) for configuration search, an AMD 9334 EPYC processor with 64 cores and 750GB memory for the scaling experiments.

\vheading{Configuration Space} We analyze \sysname's performance on a rich configuration space ($\sim$2000 points for each hardware cluster) formed by the composition of eight different configuration parameters -- mapping to different parallelization strategies and memory/compute optimizations. A summary of all the config knobs and their impact on compute system utilization is listed in \Cref{tab:knob_impacts}. All baseline systems do not support every optimization parameter shown in \Cref{tab:features_systems} and we skip these unsupported configs. Furthermore, we omit \calc and \amped baselines for the Volta architecture because they do not support modeling bfloat16.

\vspace{-1em}

\subsection{Prediction Quality}
\label{sec:eval:fidelity}

We first evaluate \sysname's accuracy in predicting the end-to-end runtime of training workloads across various models, configurations, and deployment setups. We compare \sysname against state-of-the-art: \prot, \calc, and \amped.

\vheading{Accuracy Across Configurations} \Cref{fig:eval:fidelity:pointcloud} illustrates the prediction quality of \sysname across the top one hundred configurations for training GPT3 models on four different deployment setups. \sysname consistently predicts the end-to-end runtime with high fidelity across all configurations and deployment setups. While \prot achieves comparable fidelity on V100 GPUs, it only supports a subset of configuration knobs, limiting its ability to identify top-performing configurations. Moreover, \prot's performance degrades significantly on H100 GPUs, with predictions often deviating by an order of magnitude. This is particularly surprising since \prot performs explicit profiling of kernel execution times on actual GPUs as opposed to all the other systems considered in this experiment. As shown in \Cref{fig:eval:fidelitycdf}, \amped \footnote{We contacted the authors of \prot and \amped to resolve these anomalies but could not arrive at a resolution.} consistently overestimates execution time by 2-3$\times$. Despite \calc and \amped being specialized for GPT3 training with Megatron-LM, they exhibit significantly higher prediction error compared to \sysname. \sysname achieves remarkable fidelity, predicting runtimes within 1\% error margin for $\sim$65\% of configurations on 8 V100 GPUs. This extends to larger deployments, with \sysname maintaining 10\% error margin for $\sim$90\% of configurations even at 64 H100s.

\input{figures-tex/error_cdf}

\vheading{Impact on Configuration Selection} \Cref{fig:eval:fidelity:cost} demonstrates how prediction accuracy directly impacts the identification of optimal training configurations. The graph shows the normalized cost (relative to the optimal configuration) of the best configuration selected by each system on actual deployment. \sysname consistently identifies configurations within 2\% of optimal cost across all scenarios, showcasing its ability to effectively navigate complex configuration spaces. In contrast, \prot selects configurations 5-17\% more costly than optimal, with the gap widening for larger models and GPU counts. \calc's consistent underestimation leads to configurations with 10-15\% higher costs, while \amped's overestimation results in configurations up to 56\% more expensive than optimal.
\sysname's exceptional prediction accuracy across diverse model sizes, GPU configurations, and optimization strategies directly translates to identifying highly efficient training configurations, enabling significant savings in computational resources and associated costs for large-scale deep learning training workloads.

\vheading{Breakdown of Prediction Error} The end-to-end prediction error can broadly be attributed to i) the prediction error of individual kernel runtimes, and ii) loss of detail in the emulation and simulation phases. To better characterize these errors, we compare against an \textit{oracle prediction} --- this is a modified version of \sysname that uses profiled (actual) per-kernel runtimes instead of predicted values from a regressor. The results obtained on a single-node and multi-node V100 setup are summarized in Table \ref{tab:oracle_fidelity}. 
\input{tables/v100_oracle_fidelity}

We observe that the oracle predictions are highly accurate, falling within 2\% of actual runtime in most cases, while end-to-end error is within 5-6\%. This holds across model sizes, batch sizes and arbitrary parallelism configurations --- demonstrating the importance of capturing detailed traces from emulation and validating the downstream modeling in the simulator.

\vheading{Framework Generality}
To verify the generality of \sysname's emulation approach, we test the system using training scripts scraped from popular open-source frameworks \cite{deepspeedvisiongithub, hfaccelerate}. Across a wide range of common optimization techniques and model architectures, we find that \sysname's emulation approach runs and produces traces, notably including memory optimization techniques such as ZeRO sharding and CPU offloading which involve host-device transfers (Table \ref{tab:framework_generality}). These \texttt{cudaMemCpy} operations are treated as separate kernels in \sysname, and while the offloaded tensors contain random data, the shapes collected in the trace are faithful to the actual transfer and can be used to make predictions.
\input{tables/framework_generality}

A key observation from running training scripts in the wild is that extra verification steps can occasionally lead to emulation failures if not disabled --- this is because they attempt to load and check the contents of certain portions of output buffers. We found this to be mitigated by allowing the emulator to \texttt{memcpy} small buffers to mock host-host and host-device transfers, passing most verification checks that inspect metadata (such as a tensor count or rank order).

To further validate the efficacy of \sysname across model architectures, we collect results from a representative vision model, ResNet152 (\Cref{fig:eval:resnet152}) on an 8xA40 node. This specific workload is particularly challenging due to heterogeneous GPU links and the use of \texttt{torch.compile} (Appendix \ref{appendix:predictions}). Despite this, we observe consistent high-fidelity runtime predictions with less than 5\% error over half of all configurations, similar to our experiments with Megatron-LM.

\input{figures-tex/resnet_eager_scaling}

\input{tables/config_search_space}

\subsection{Configuration Search with \sysname}
\label{sec:eval:opti}

We ran a hyperparameter search using our system over the Megatron-LM configuration space for each resource/model specification (\Cref{tab:eval-config-knobs}). The system was configured to use CMA-ES~\cite{cma, hansen2001completely} as the search algorithm. Further, we enabled all of our optimizations: dynamic worker de-duplication, inter+intra trial concurrency, and fidelity-preserving trial pruning (using Megatron-LM specific tactics, detailed in Appendix \ref{appendix:tactics}). The early stopping mechanism was configured to terminate the search if the MFU of the top 5 configs remained the same for 20 consecutive non-OOMing configs.

\input{figures-tex/e2e-config-search}
\vheading{End to end performance} The search completed in under an hour across all resource/model specs (\Cref{fig:config-search-runtime}). %
Further, the search was able to find configurations very close to if not the same as the optimal across all resource specs (\Cref{fig:config-search-mfu-optimality}).

\subsection{Supporting Hyperscale Workloads}
\label{sec:eval:hyperscale}

Our experiments thus far have maintained transparency, requiring no domain-specific knowledge of the workload. This also applies to worker deduplication (\Cref{sec:deduplication}) --- in order to identify which workers are duplicates, the system first emulates all workers for at least one iteration. This presents a challenge when attempting to scale \sysname to large clusters with thousands of GPUs.

With some explicit knowledge of the workload, however, we observe that unique workers can be identified ahead of time. For instance, in Megatron-LM, we can calculate which ranks would participate in tensor, data, and pipeline communication using the parallelism configuration (\cref{tab:eval-config-knobs}). This determines the set of unique workers --- specifically, the first data-parallel rank of each communicator group and every pipeline parallel rank. Using this information, we extend \sysname to \textit{selectively launch} unique workers, drastically reducing overheads. 

This optimization enables us to study the behavior of clusters with up to 16K GPUs. Since we did not have access to clusters of this size for profiling collectives, we integrated with ASTRA-sim \cite{won2023astra} for network simulation. First, keeping the parallelism configuration fixed (TP8, PP8, 12K batch size, 64 microbatches), we vary the data-parallel degree (\Cref{fig:eval:h100_dp_scaling_mfu}). The results demonstrate the expected trend of \textbf{\textit{sublinear scaling}} --- as the number of GPUs is scaled, communication overhead dominates and leads to low MFU.
\input{figures-tex/h100_dp_scaling}

In \cref{fig:eval:eager-scaling}, we keep the configuration entirely fixed and scale the global batch size. The largest configuration takes $\sim$25 minutes to run using 8 unique workers, each corresponding to a pipeline parallel rank. While not conducive to an exhaustive config search, these results demonstrate that \sysname can effectively scale to thousands of GPUs.

\subsection{Ablation studies}
\label{sec:eval:ablation}

\vheading{Impact of dynamic worker deduplication} To quantify the impact of dynamic worker deduplication on \sysname's end-to-end runtime, we fix the parallelism configuration and increase the data parallel degree (thereby testing a larger cluster). Any new DP workers added would be redundant from the perspective of emulation --- this allows us to isolate the impact of dynamic worker deduplication. \Cref{fig:ablation-unique-ranks} illustrates the results.
\input{figures-tex/dedup_trial_skipping}

Without worker deduplication, we observe a significant increase in runtime, with the H100 64 GPU run taking approximately two hours. This is because the system has to emulate and subsequently simulate the execution of every GPU, which results in increased overhead as the number of GPUs increases. In contrast, with dynamic worker deduplication, we observe that the runtime remains approximately the same, with the H100 64 GPU run now taking only 7 minutes -- a 94\% improvement. We attribute this to the following. First, deduplication eliminates both the emulation and simulation of redundant GPUs. Second, scaling certain parallelism configuration knobs does not impact the number of \textit{unique} workers; this can be exploited to improve efficiency.

\vheading{Impact of fidelity-preserving trial pruning} For the configuration search carried out in \Cref{sec:eval:opti}, the trial skipping mechanism skipped around 20-30\% of configurations (\Cref{fig:config-search-heuristic-performance}) across all resource/model specs, thereby playing a considerable role in bringing down the overall search time.

\input{tables/h100-32-e2e-ablation}
\vheading{Impact of optimizations on config search runtime} To evaluate the impact of all optimizations (including the use of the CMA search algorithm) on overall search runtime, we compare against grid search without any heuristic optimizations. As evident in \Cref{tab:eval-h100-32-ablation}, the optimizations significantly reduce the overall search time, bringing it down from over a day to just under 40 minutes. Worker deduplication is a key enabler for this reduction since it reduces the resource usage of each trial, enabling greater concurrency. This is corroborated by the increased OOM rate when running the full set of workers without optimizations. \sysname's applicability to large configuration spaces would not be possible without deduplication and trial pruning.

%% file: figures-tex/relative_cost.tex
\begin{figure*}[htbp]
   \centering
   \includegraphics[width=0.8\textwidth]{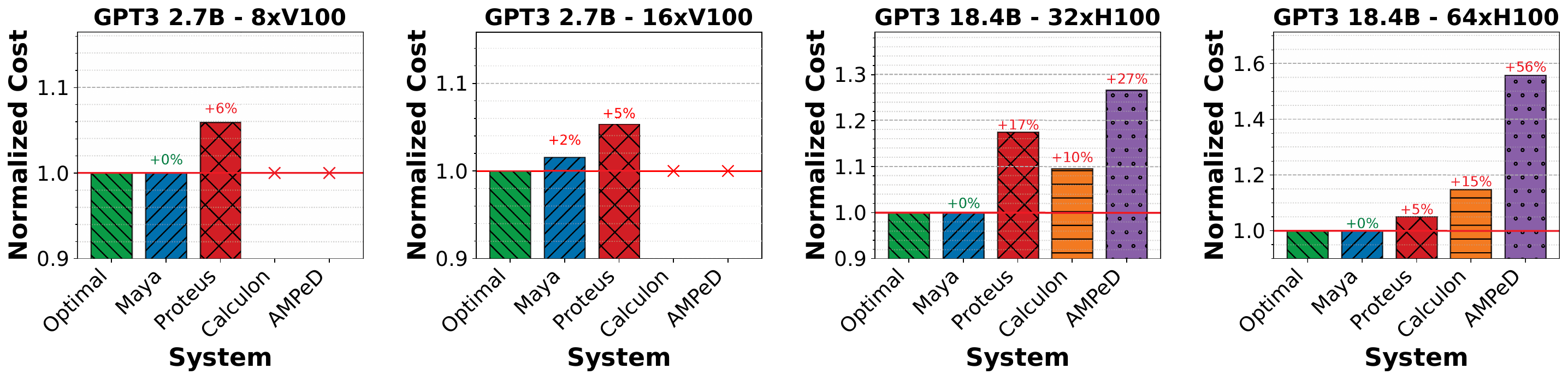}
   \caption{Cost impact of prediction accuracy on configuration selection. We evaluate GPT3-2.7B and GPT3-18.4B models across V100 and H100 clusters, showing the cost of each system's selected configuration normalized to the optimal configuration's cost. \sysname consistently identifies configurations within 2\% of optimal cost, while baseline systems can result in up to 56\% higher cost. \todo{regen the plot with maya name}}
   \label{fig:eval:fidelity:cost}
\end{figure*}

%% file: figures-tex/error_cdf.tex
\begin{figure}[htbp]
    \centering
    \includegraphics[width=\linewidth]{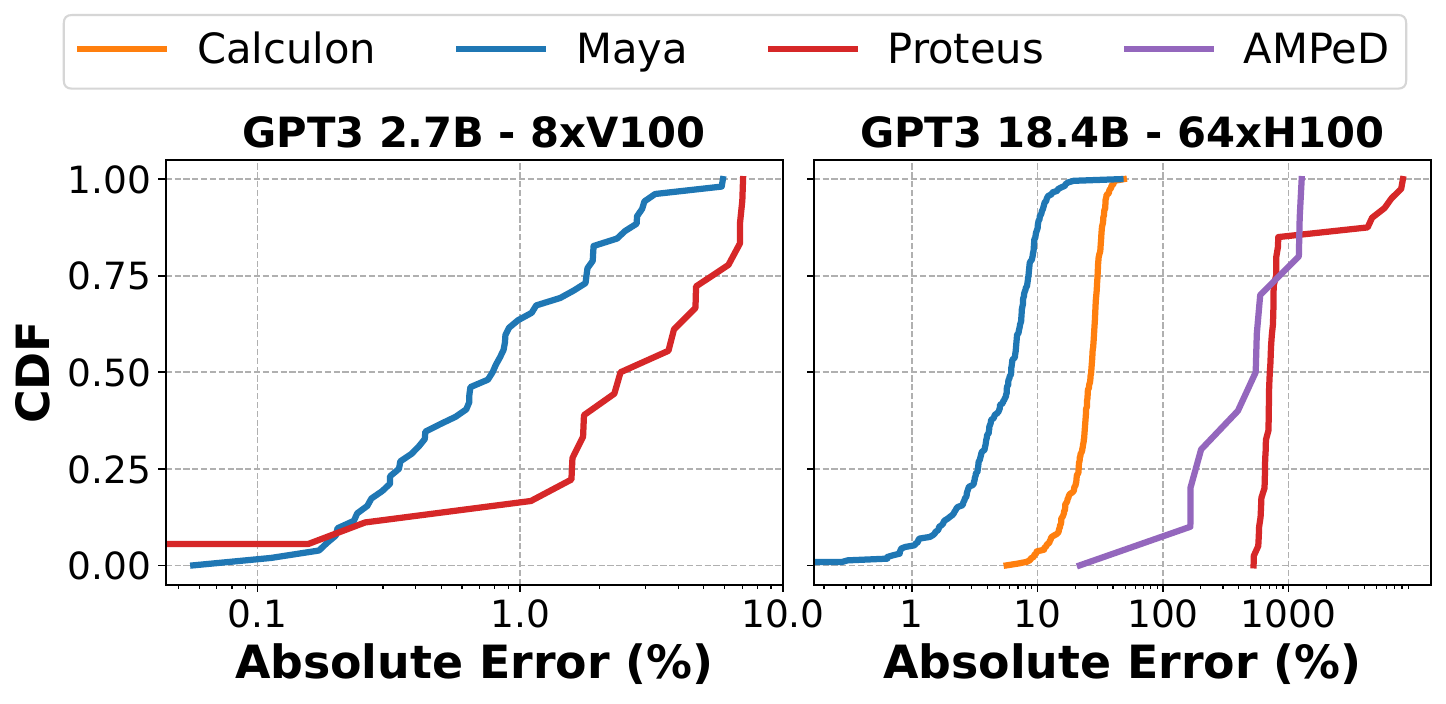}
    \caption{Cumulative distribution of prediction errors across configurations. \sysname achieves less than 1\% prediction error for 65\% of configurations on 8{\myx}V100 cluster. \sysname achieves sub 10\% prediction error for 90\% of configurations on 64{\myx}H100 cluster, while baseline systems show 10-1000\% errors. \todo{regen the plot with maya name}}
    \label{fig:eval:fidelitycdf}
\end{figure}

%% file: tables/v100_oracle_fidelity.tex
\begin{table}[htbp]
\vspace{1em}
\centering
\begin{tabular}{p{1.8cm}cccccc}
\toprule
\textbf{Model} & \textbf{BS} & \textbf{TP} & \textbf{PP} & \textbf{GA} & \textbf{Oracle} (\%) & \textbf{E2E} (\%) \\
\midrule
\multirow{5}{*}{\parbox{1.8cm}{GPT3-1.3B \newline (8 GPUs)}} 
  & 16 & 1 & 2 & 2 & 0.60 & 1.80 \\
  & 16 & 2 & 1 & 2 & 1.00 & 3.60 \\
  & 16 & 2 & 2 & 2 & 1.20 & 2.20 \\
  & 16 & 2 & 4 & 2 & 0.50 & 2.60 \\
  & 16 & 4 & 2 & 2 & 4.10 & 3.20 \\
\midrule
\multirow{5}{*}{\parbox{1.8cm}{GPT3-2.7B \newline (8 GPUs)}} 
  & 16 & 1 & 2 & 2 & 0.70 & 0.30 \\
  & 16 & 2 & 1 & 2 & 2.70 & 6.50 \\
  & 8  & 2 & 2 & 2 & 0.60 & 5.00 \\
  & 8  & 2 & 4 & 2 & 0.14 & 3.50 \\
  & 8  & 4 & 2 & 2 & 6.00 & 4.00 \\
\midrule
\multirow{5}{*}{\parbox{1.8cm}{Llama2-7B \newline (32 GPUs) \newline}} 
  & 16 & 2 & 8 & 2 & 0.15 & 0.40 \\
  & 8 & 2 & 8 & 4 & 0.80 & 1.80 \\
  & 16 & 4 & 4 & 2 & 3.10 & 1.40 \\
  & 8  & 8 & 2 & 2 & 1.01 & 1.09 \\
\bottomrule
\end{tabular}
\caption{Breakdown of error on V100 with varying batch size (BS), tensor parallelism (TP), pipeline parallelism (PP) and gradient accumulation (GA). \textit{Oracle} represents a modified version of our system that uses oracular (i.e. actual) kernel runtimes --- illustrating the error introduced by the device emulation and simulation phases. \textit{E2E} captures the end-to-end error including errors from kernel-level mispredictions. \textit{Oracle} predictions are closer to the actual runtime than \textit{E2E}, barring a few cases attributable to noise.}
\label{tab:oracle_fidelity}
\end{table}

%% file: tables/framework_generality.tex
\begin{table}[htbp]
\vspace{1em}
\begin{tabular}{m{0.2\linewidth} m{0.3\linewidth} p{0.4\linewidth}}
    \toprule
    \textbf{Framework} & \textbf{Optimizations} & \textbf{Models} \\
    \midrule
    DeepSpeed & ZeRO 1-3, \newline Act. Offload & \multirow{3}{\linewidth}{ResNet, DenseNet, MobileNet, VGG, BERT, GPT, Llama, T5, ViT} \\
    \cmidrule{1-2}
    PyTorch & \texttt{torch.compile}, FSDP, DDP \\
    \bottomrule
\end{tabular}
\caption{Frameworks and models tested with \sysname emulation, in addition to Megatron-LM.}
\label{tab:framework_generality}
\end{table}

%% file: figures-tex/resnet_eager_scaling.tex
\begin{figure}[!hbt]
    \centering
    \includegraphics[width=0.5\linewidth,keepaspectratio]{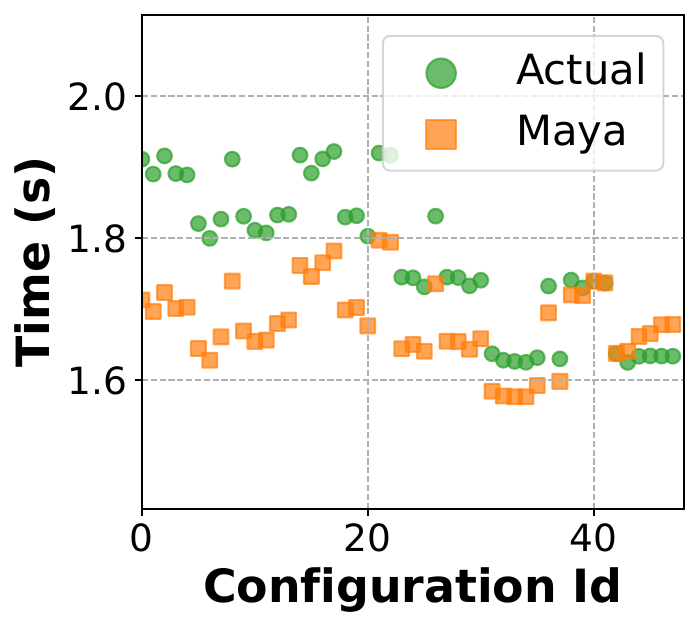}
    \caption{Prediction accuracy of \sysname across different configurations of ResNet152 deployed on 8xA40 GPUs.}
    \label{fig:eval:resnet152}
\end{figure}

%% file: tables/config_search_space.tex
\begin{table}[htbp]
\vspace{1em}
\centering
\begin{tabular}{l r}
\toprule
\textbf{Configuration Knob} & \textbf{Search Space} \\
\midrule
Tensor Parallel Degree & 1, 2, 4, 8 \\
Pipeline Parallel Degree & 1, 2, 4, 8 \\
Microbatch Multiplier & 1, 2, 4, 6, 8 \\
Number of Virtual Stages & 1, 2, 4 \\
Activation Recomputation & True, False \\
Sequence Parallelism & True, False \\
Distributed Optimizer & True, False \\
\bottomrule
\end{tabular}
\caption{Configuration knobs and their search space.}
\label{tab:eval-config-knobs}
\end{table}

%% file: figures-tex/e2e-config-search.tex
\begin{figure}[htbp]
\vspace{1em}
    \begin{subfigure}[h]{0.235\textwidth}        
        \includegraphics[width=\textwidth]{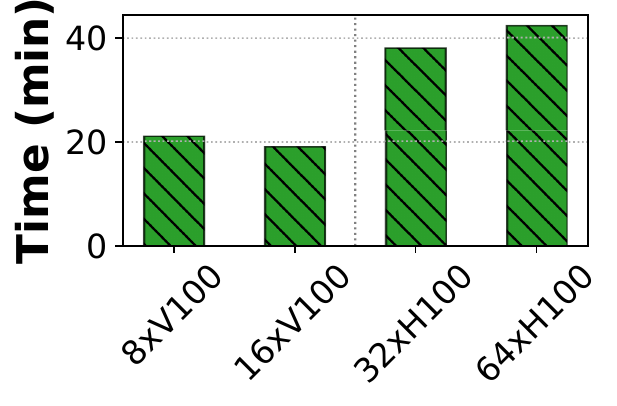}
        \caption{Configuration search runtime}
        \label{fig:config-search-runtime}
    \end{subfigure}
    \begin{subfigure}[h]{0.235\textwidth}        \includegraphics[width=\textwidth]{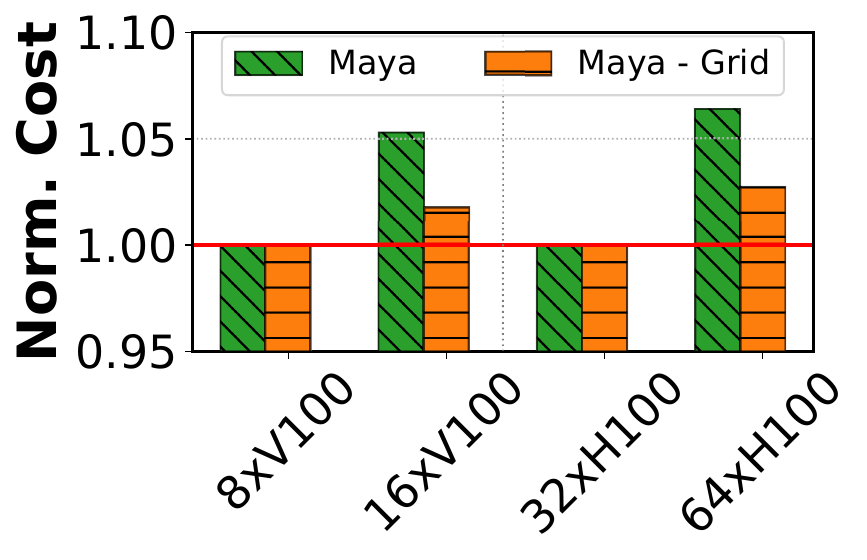}
        \caption{Normalized cost}
        \label{fig:config-search-mfu-optimality}
    \end{subfigure}
    \caption{End-to-end runtime and fidelity of configuration search. We compare the normalized cost of configurations found using \sysname against the optimal. For reference, we also include the optimal configuration found using grid search with \sysname.}
\end{figure}

%% file: figures-tex/h100_dp_scaling.tex
\begin{figure}[htbp]
    \centering
    \begin{minipage}[t]{0.48\linewidth}
        \centering
        \includegraphics[width=0.9\textwidth,keepaspectratio]{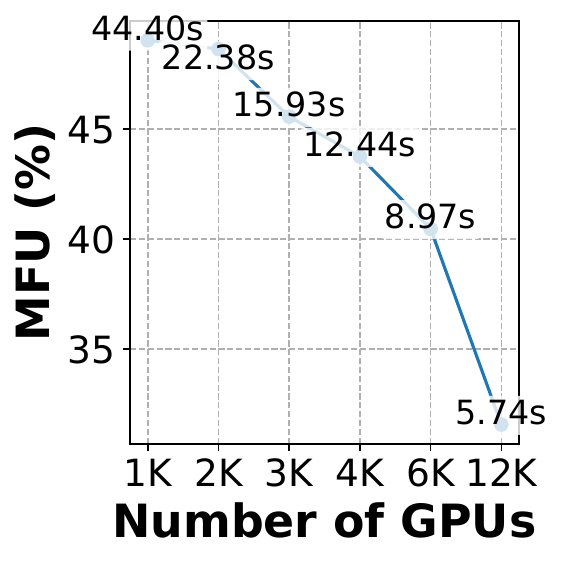}
        \caption{Predicted MFU and iteration times when scaling DP.}
        \label{fig:eval:h100_dp_scaling_mfu}
    \end{minipage}
    \hfill
    \begin{minipage}[t]{0.48\linewidth}
        \centering
        \includegraphics[width=0.92\textwidth,keepaspectratio]{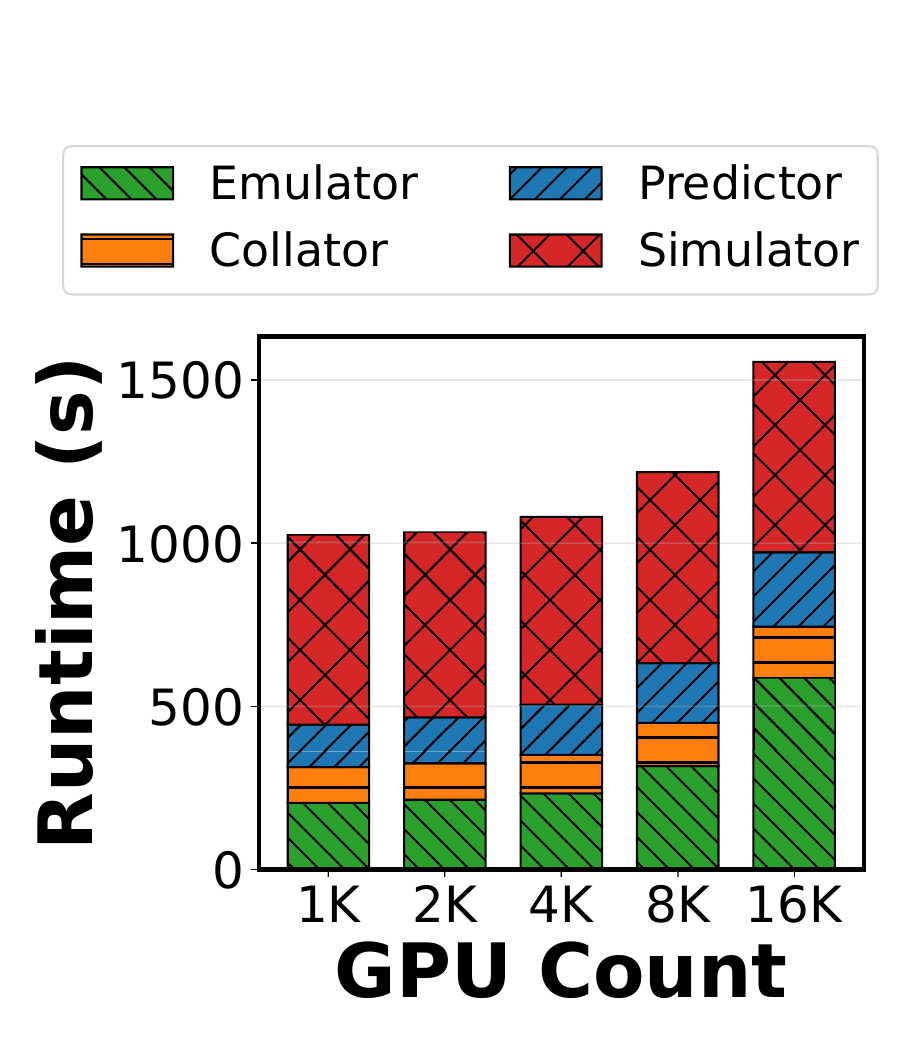}
        \caption{\sysname stack runtime when scaling to 16K GPUs.}
        \label{fig:eval:eager-scaling}
    \end{minipage}
\end{figure}

%% file: figures-tex/dedup_trial_skipping.tex
\begin{figure}[htbp]
    \centering
    \begin{minipage}{0.48\linewidth}
    \centering
    \includegraphics[width=\linewidth]{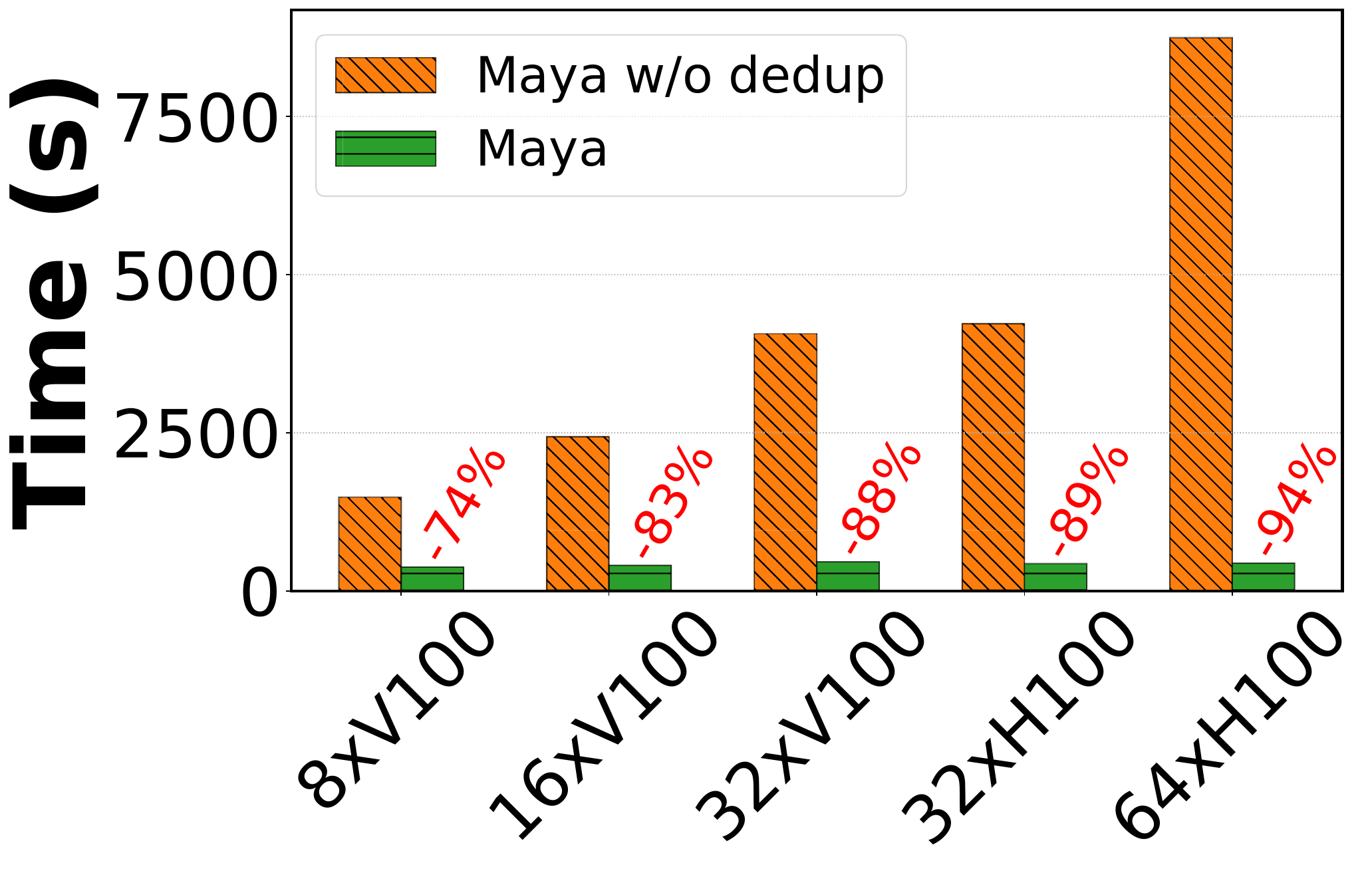}
\caption{Impact of worker deduplication on runtime.}
    \label{fig:ablation-unique-ranks}
    \end{minipage}
    \hfill
    \begin{minipage}{0.48\linewidth}
    \centering
    \includegraphics[width=\linewidth]{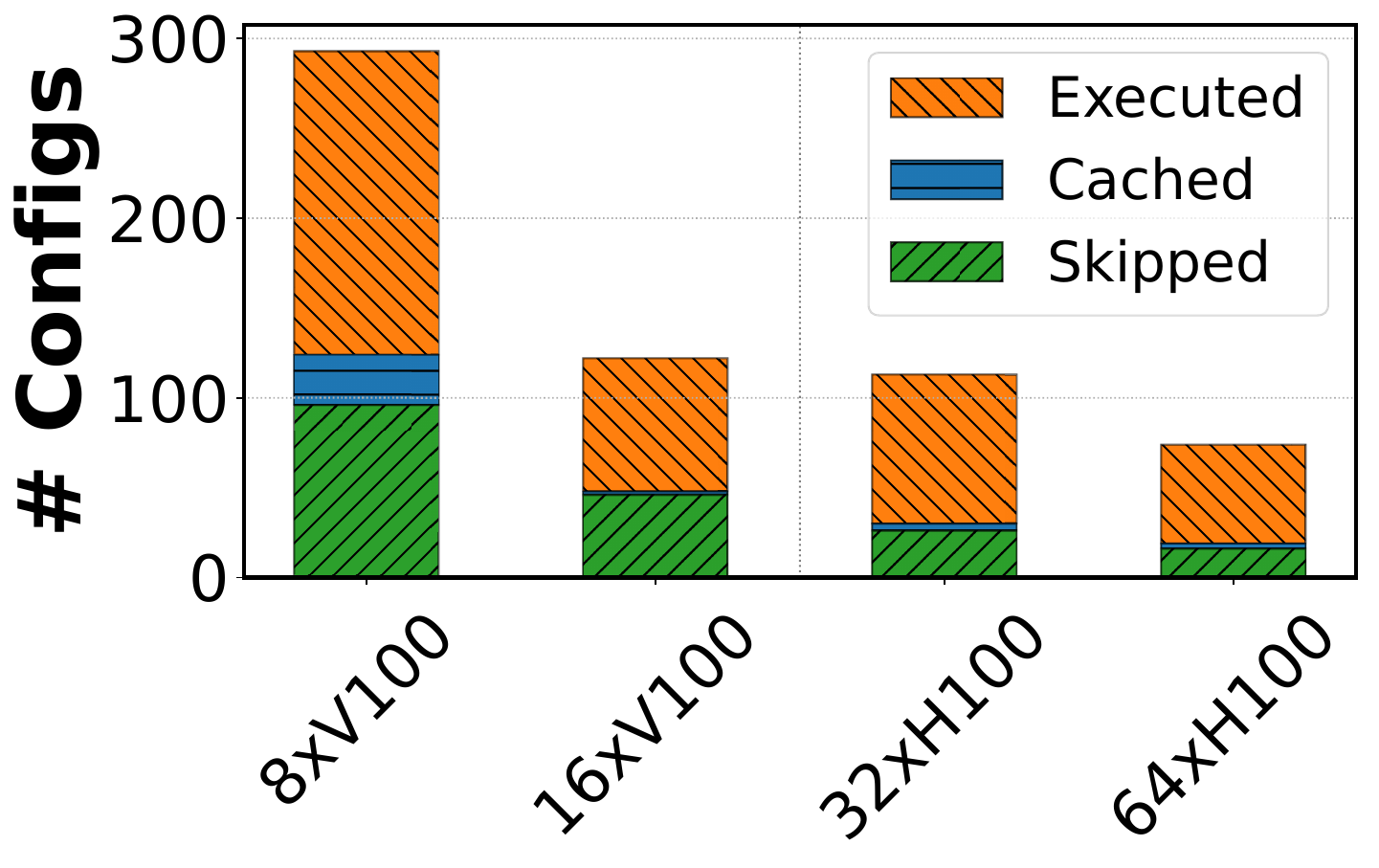}
    \caption{Trial status breakdown during config search.}
    \label{fig:config-search-heuristic-performance}
    \end{minipage}
\end{figure}

%% file: tables/h100-32-e2e-ablation.tex
\begin{table}[htbp]
\vspace{1em}
\centering
\begin{tabular}{c c c}
\toprule
\textbf{Stage} & \sysname & No Optimization \\
\midrule
Emulation & 9m & 14m \\
Trace Collation & 2m & 7m \\
Runtime prediction & 1.5m & 8m \\
Simulation & 4.5m & 55m \\
\midrule
Total search time & 38m & >24hrs\\
\bottomrule
\end{tabular}
\caption{Runtime statistics of configuration search on the H100 32 resource/model spec with and without optimizations enabled. The per-stage times are averaged across all trials.}
\label{tab:eval-h100-32-ablation}
\end{table}

%% file: 8-discussion.tex
\section{Discussion}
\vheading{Taxonomy of CPU computation} \sysname models host-side overheads as wall-clock time measurements between API calls to the emulator. This allows arbitrary host logic to be abstracted away, while still accounting for the impact of these overheads on end-to-end latency. However, there are workloads where significant CPU computation is involved, and this could affect prediction accuracy if there are hardware differences between the machine used during emulation vs. the target cluster. This can be addressed by applying the per-operation prediction approach to CPU work instead of simply collecting a wall-clock time, though this may not be exposed through a narrow API surface like accelerators. A combination of these two models could enable more general CPU overhead estimates.

\vheading{Dynamic control flow} As a result of relying on emulation, \sysname does not model computation graphs where the control flow depends on the result of tensor computation. This assumption is shared with several DL compilers and parallelism search engines \cite{zheng2022alpa}, \cite{flexflow}. Mixture-of-Experts (MoE) architectures display this pattern --- while most expert-parallel kernels used for MoE training \cite{deepep2025, pplxrepo} remove the need for data-dependent control-flow, there are some implementations that use gating on the host.

For expert-parallel kernels, runtime predictors can be trained by encoding the input distribution during the profiling phase \cite{lin2025apexextensibledynamismawaresimulator, vidur}, keeping the rest of the \sysname flow unchanged. To handle host-side gating, annotations on the source model can be used to identify the gating function --- instead of returning a random tensor during emulation, we would sample a distribution to generate a spread of runtimes.

\vheading{SM Contention} \sysname assumes decoupling between network collectives and concurrent compute streams. As a result, while we are able to model overlapping streams and arbitrary synchronization, we cannot trivially model SM-level interference where network and compute kernels contend for resources. It could be possible to modify and extend the simulator to identify such patterns and scale predicted durations accordingly --- we leave this to future work.

%% file: 9-related.tex
\section{Related Work}

The growing computational demands of training large foundation models have driven significant research into performance modeling and optimization of DLT workloads. 

\vheading{Kernel runtime prediction} Habitat~\cite{geoffrey2021habitat} extrapolates single-GPU measurements to predict cross-device performance. More recent approaches such as NeuSight \cite{neusight}, Omniwise \cite{omniwise} rely on a mix of profiling data and architectural details of the accelerator to predict kernel runtimes more accurately. ASTRA-sim~\cite{rashidi2020astra} focuses specifically on network topology and collective communication modeling. These are complementary to \sysname and can be plugged in as needed, enabling end-to-end estimates on a wide range of workloads.

\vheading{Analytical Performance Models} Analytical models predict DLT performance through mathematical formulations of system behavior. AMPed~\cite{moolchandani2023amped} and Calculon~\cite{isaev2023calculon} propose specialized models for LLMs but support only limited parallelization strategies and require explicit modeling of new optimizations. Other work focuses on specific architectures like CNNs~\cite{yan2015performance,qi2017paleo,gianniti2018performance}. While these techniques can provide quick estimates, their applicability is limited to specific models and configurations.

\vheading{Domain-Specific Simulators} Simulation-based approaches aim to capture detailed system behavior through explicit modeling. Proteus~\cite{duan2023proteus} introduces a strategy tree abstraction for modeling parallelization patterns but requires translation into a custom specification language. DistIR~\cite{santhanam2021distir} proposes an intermediate representation for distributed computations but struggles with complex parallelization strategies. Daydream~\cite{zhu2020daydream} captures dependency graphs from execution traces, but requires GPU access and manual optimization modeling. vTrain~\cite{bang2023vtrain} uses CUPTI profiling to measure kernel runtimes but faces challenges modeling communication patterns in complex parallelization strategies.

%% file: 10-conc.tex
\section{Conclusion}
\label{sec:conclusion}
Training large foundation models at scale has made the optimization of training recipes for hardware utilization a critical challenge, with costs reaching hundreds of millions of dollars. We introduce, \sysname, a runtime modeling system, addresses this challenge through a fundamental insight: by operating at the narrow interface between training frameworks and accelerator devices, we can eliminate the semantic gap that forces existing systems to trade off between accuracy, usability, and generality. Through transparent device emulation and precise runtime simulation, \sysname achieves prediction accuracy within \evalpe\% error and identifies configurations within \evalcg\% of optimal cost across diverse scenarios, from small V100 clusters to large-scale H100 deployments. As distributed training continues to push the boundaries of scale and complexity, \sysname's transparent runtime modeling approach is a crucial step toward sustainable and efficient deployment of large-scale AI systems.

%% file: 11-ack.tex
\section*{Acknowledgments}
This material is based on work that was partially supported by the National Science Foundation under grant number CNS-2420977. We would like to express our sincere gratitude to the reviewers, the PC panel, and especially our shepherd Prof. Richard Mortier for their insightful comments and thoughtful consideration, which significantly improved the quality of this paper. \\
\textbf{Disclaimer}: Any opinions, findings,
conclusions, or recommendations expressed in this material are those of the authors and do not necessarily
reflect the views of the National Science Foundation.

%% file: 11-appendix.tex
\appendix

\section{Key Algorithms}
\label{appendix:simulator_core}

\vheading{Simulator}
The core logic of the simulator is summarized in Algorithm \ref{alg:simulator}. At a high-level, the simulator handles discrete events and increments a clock on the completion of each event. Events are classified into different types based on their effect --- for instance, some events indicate the enqueuing of a kernel on the GPU, while others indicate a synchronization between device streams.

The scheduler (Algorithm \ref{alg:scheduler}) handles updating the state of each host and device in the simulated cluster and adding or removing operations from the corresponding queues. Resources that are busy will cause any new ops targeting them to be queued, effectively simulating blocking delays by deferring their execution. Every time an operation completes execution, the next scheduler tick pops an operation from the resource-specific queue and adds an $\texttt{EndEvent}$ marking its completion in the future. Every discrete $\texttt{EndEvent}$ in the top-level queue is followed by a scheduler tick, ensuring that operations do not block forever on resources.
\input{pseudocode/simulator}

\input{pseudocode/waitmaps}
\vfill\break
The simulator maintains two global structures to track synchronization across hosts and accelerators --- the CUDA Event Wait Map, and the Network Collective Wait Map. These are detailed in Algorithm \ref{alg:waitmaps}. Corresponding event types that perform synchronization lookup these maps in their handlers. For example, in the case of $\texttt{cudaEventSynchronize}$, the specific device stream or host that intends to block while waiting for a CUDA event inserts an entry in the wait-map (keeping its associated resources from processing ops), while the matching $\texttt{cudaEventRecord}$ triggers the resources to be freed by creating $\texttt{EndEvent}$ instances for the blocking ops.

A similar mechanism is used to model collectives --- each worker makes an entry in the collective wait map. Once the final worker of the collective joins, all the corresponding streams are unblocked and can proceed. In this case, $\texttt{EndEvent}$s are scheduled with a timestamp after the predicted duration of the collective; effectively, \sysname models the delays involved in starting a collective using a global sync point and then assumes that workers move in lockstep. Any effects associated with the on-the-wire time of the collective can thus be abstracted away in the predicted time --- while this is not completely faithful to the setup/teardown of NCCL collectives, it is sufficiently accurate for an end-to-end accounting of latency.

These fairly simple structures can express a wide variety of possible synchronization behaviors since they operate at the CUDA stream level. Computation streams can overlap with collectives since each stream is a separate resource. The host queue can block on a specific CUDA event or a device stream, deferring the execution of future CUDA API calls. An arbitrary pipeline parallel schedule is a combination of such synchronizing events, and thus \sysname can trivially capture these behaviors without any explicit modeling.

\section{Per-kernel prediction accuracy}
\label{appendix:predictions}
The default predictors in \sysname use random forest regressors trained on per-kernel runtime data. Tables \ref{tab:h100_kernel_accuracy}, \ref{tab:v100_kernel_accuracy} and \ref{tab:a40_kernel_accuracy} include metrics on the prediction error of the kernels trained for Megatron-LM (H100, V100) and PyTorch FSDP (A40). All results use a random 80:20 training/test data split.

As a general theme, we observe the same characteristics as \cite{geoffrey2021habitat}, \cite{zhu2020daydream} --- a small portion of the kernels are responsible for a significant portion of end-to-end prediction error (matmuls for language models, convolution kernels for vision models). As a result, even large percentage-wise errors in several other kernels do not cause any significant degradation in end-to-end accuracy.

In keeping with this observation, we conduct more extensive profiling of these heavy-hitter kernels --- sweeping a large space of input dimensions for convolution/matmul. The remaining kernels are scraped from traces, collecting by running a single-layer LLaMa/OPT/vision model over a range of batch sizes and tensor-parallel dimensions (since other optimizations like pipeline parallelism do not affect the runtime of a single kernel). The training set for the heavy-hitter kernels included $\approx$ 42k individual points, compared to a few thousand points each for the rest.

\input{tables/h100_kernel_accuracy}
\input{tables/v100_kernel_accuracy}

In contrast to computation operations, there is a much smaller set of network collectives (<10) that is used in deep learning workloads. Furthermore, the input space of these operators typically much smaller, typically comprising only two parameters -- number of workers and input size. This allows us to devise a simple policy for modeling these operations. We first  collect performance data in a fashion similar to \texttt{nccl-tests}. We only sample data in the range that is generally relevant for training workloads ranging from tens of megabytes to tens of gigabytes. We then use our regression pipeline to interpolate within this range. While this affects generalization to dimensions outside the range of the training set, this does not pose a problem in practice since the collective sizes are bounded by the batch size, model parameters and accelerator memory. %

Automatically generated fused kernels pose a unique challenge due to an explosion in generated kernel signatures --- arising from a large number of op combinations. We address this by collecting information from the compiler IR about the content of the kernels rather than just their inputs. In our experiments with the compiler-fused Triton kernels used in \texttt{torch.compile}, features such as the number of primitive Triton language instructions (add, sub etc.) in the kernel definition proved valuable in predicting kernel runtimes. The corresponding training data was collected by sweeping workload traces for different models/batch sizes and extracting the relevant features/runtimes. Through this approach, we achieve comparable accuracy to that of the kernel predictions trained on focused micro-benchmarks. 

\vspace{1em}

\input{tables/a40_kernel_accuracy}

\section{Performance of alternate search algorithms}

Tune supports several search algorithms out of the box. We investigated the performance of a subset of these algorithms by the progress of the search at distinct phases. Each phase was defined by the number of unique valid configurations sampled by the algorithm up to that point. Every algorithm (with the exception of grid search) was allocated a budget of 2000 samples. \Cref{fig:config-search-alg-comparison} shows the results of this experiment, where the MFU is computed from the iteration times predicted by \sysname. Interestingly, despite the fact that these algorithms are general-purpose and therefore lack domain-specific knowledge of the search space, they appear to converge after having sampled about 200 to 300 unique valid configurations, a 60-75\% improvement over grid search.

\section{Fidelity-preserving Tactics for the Megatron-LM Search Space}
\label{appendix:tactics}

We leverage the performance characteristics of certain Megatron-LM configuration knobs to devise four fidelity-preserving tactics, summarized in \Cref{table:fidelity-preserving-tactics}.
\input{figures-tex/search_algo_comparison}

\input{tables/config_search_optimizations}

%% file: pseudocode/simulator.tex
\begin{algorithm}
\caption{Core Discrete-Event Simulator Algorithm}
\label{alg:simulator}
\begin{algorithmic}[1]
\Procedure{Simulate}{config, host\_op\_trace}
    \State $\textit{time} \gets 0$
    \State $\textit{event\_queue} \gets \text{ PriorityQueue()}$
    \State $\textit{cluster} \gets \text{ Cluster(config)}$
    \State $\textit{scheduler} \gets \text{ Scheduler(config, cluster)}$
    
    \Statex \LComment{Init queue with host ops and inter-host-op overheads from the input trace}
    \For{each $\textit{host\_op} \lor \textit{overhead}$  in $\textit{host\_op\_trace}$}
        \State $\textit{event} \gets \text{ HostOpArrivalEvent}(\textit{host\_op})$
        \State $\textit{overhead} \gets \text{ HostOverhead}(\textit{overhead\_dur})$
        \State $\textit{event\_queue.put(event)}$
        \State $\textit{event\_queue.put(overhead)}$

    \EndFor
    
    \Statex
    \While{$\neg \textit{event\_queue.empty()}$}
        \LComment{Get the next chronological event}
        \State $\textit{event} \gets \textit{event\_queue.get()}$
        
        \Statex \LComment{Update simulation time.}
        \State $\textit{time} \gets \textit{event.end\_time}$
        
        \Statex \Comment{Handle event polymorphically based on its type.}
        \State $\textit{new\_events} \gets \textit{event.handle\_event(scheduler)}$
        
        \LComment{Add newly generated events to the queue}
        \For{each $\textit{new\_event}$ in $\textit{new\_events}$}
            \State $\textit{event\_queue.put(new\_event)}$
        \EndFor
        
    \EndWhile
    
    \Statex
    \Return $\textit{time}$
\EndProcedure
\end{algorithmic}
\end{algorithm}

\begin{algorithm}
\caption{Scheduler Event Handling Logic}
\label{alg:scheduler}
\begin{algorithmic}[1]
\Procedure{Event.handle\_event}{scheduler}
    \State \Comment{The logic here is polymorphic, depending on the concrete event type.}
    \Statex
    
    \If{$\textit{event}$ is an \textbf{OpArrivalEvent}}
        \LComment{An operation from the trace has arrived (e.g., kernel launch).}
        \State \Return $\textit{scheduler.schedule\_operation(event.op)}$
    
    \ElsIf{$\textit{event}$ is an \textbf{EndEvent}}
        \LComment{An operation has finished, freeing a resource.}
        \State \Return $\textit{scheduler.op\_complete(event.op)}$
    \ElsIf{$\textit{event}$ is a \textbf{ScheduleEvent}}
        \LComment{A global scheduling tick occurs.}
        \State $\textit{newly\_started\_ops} \gets \textit{scheduler.schedule()}$
        \State \Return $\text{create\_end\_events}(\textit{newly\_started\_ops})$
        
    \Else
        \LComment{Handle other event types (e.g., sync, collective).}
        \State \Return $\textit{handle\_other\_events(scheduler, event)}$
    \EndIf
\EndProcedure
\end{algorithmic}

\vspace{0.5cm}

\begin{algorithmic}[1]
\Procedure{Scheduler.schedule()}{}
\State $\textit{newly\_started\_ops}$ = $\emptyset$
    \For{each $\textit{device, stream}$ in $\textit{scheduler.cluster}$ }
        \If{$\textit{device.is\_busy()} \lor \textit{stream.is\_busy()}$}
            \LComment{A required resource is busy, so don't deque}
            \State $\textbf{continue}$
        \Else
            \LComment{Resources are free, so process the op}
            \State $\textit{op} \gets \textit{device\_queue}$.front() 
            \State $\textit{device.set\_busy()}$
            \State $\textit{stream.set\_busy()}$
            
            \State $\textit{duration} \gets \text{get\_runtime(op)}$
            \State $\textit{end\_time} \gets \textit{current\_time} + \textit{duration}$
            \State $\textit{end\_event} \gets \text{ EndEvent}(\textit{end\_time}, \textit{op})$
        \EndIf
    \EndFor
    \State \Return $\textit{newly\_started\_ops}$
\EndProcedure
\end{algorithmic}

\vspace{0.5cm}

\begin{algorithmic}[1]
\Procedure{Scheduler.op\_complete}{completed\_op}
    \State $\textit{device}, \textit{stream} \gets \text{completed\_op.get\_resources()}$
    
    \LComment{Mark the resources as free}
    \State $\textit{device.set\_free()}$
    \State $\textit{stream.set\_free()}$
    
    \If{$\textit{device.wait\_queue.is\_not\_empty()}$}
        \State \Comment{Check for and schedule the next pending operation}
        \State $\textit{next\_op} \gets \textit{device.wait\_queue.get\_next()}$
        \State \Return $\text{schedule\_operation}(\textit{next\_op})$
    \Else
        \State \Return $\emptyset$ \Comment{No pending work for this resource}
    \EndIf
\EndProcedure
\end{algorithmic}
\end{algorithm}

%% file: pseudocode/waitmaps.tex
\begin{algorithm}
\caption{Synchronization Wait Map Structures}
\label{alg:waitmaps}
\begin{algorithmic}[1]
\algblock{CudaEventWaitMap}{EndCudaEventWaitMap}
\CudaEventWaitMap
    \State \Comment{\textbf{Structure:} Map from a CUDA (event ID, version) pair to a list ops waiting for it. Versions track re-use of the same CUDA event handle.}
    \State $\textit{events}: (\textit{event\_id}, \textit{version}) \to \textit{waiting\_ops}$
    \Statex
    \Procedure{BlockOnEvent}{event\_id, version, op}
        \LComment{An operation `op` (from a host or stream) blocks on a future event.}
        \State $\textit{events}[\textit{event\_id},\textit{version}].add(op)$
        \State Stall the host/stream associated with $\textit{op}$
    \EndProcedure
    \Statex
    \Procedure{ReleaseWaiters}{event\_id, version}
        \State \Comment{The event has been recorded; release all waiting operations for scheduling.}
        \State $\textit{released\_ops} \gets \textit{events}.pop(\textit{event\_id},\textit{version})$
        \For{each $op$ in $released\_ops$}
        \State Free the host/stream associated with $op$ by creating the associated $EndEvent$ instances
        \EndFor
        \State \Return $\textit{released\_ops}$
    \EndProcedure
\EndCudaEventWaitMap
\end{algorithmic}

\vspace{0.5cm}
\hrule
\vspace{0.5cm}
\begin{algorithmic}[1]
\algblock{NetworkCollectiveWaitMap}{EndNetworkCollectiveWaitMap}
\NetworkCollectiveWaitMap
    \State \Comment{\textbf{Structure:} A map from a NCCL collective's unique ID to its list of participant kernels.}
    \State $\textit{collectives}: (\textit{nccl\_group\_id}, \textit{call\_idx}) \to \textit{kernels}$
    \Statex
    \Procedure{JoinCollective}{kernel}
        \State \Comment{A device's kernel joins a collective operation and waits for peers.}
        \State $collectives[\textit{group\_id}, \textit{call\_idx}].add(kernel)$
        \State $\textit{wait\_list} \gets collectives[\textit{group\_id}, \textit{call\_idx}]$
        \Statex
        \If{$\text{length}(\textit{wait\_list}) = \textit{kernel.num\_ranks}$}
            \LComment{The last worker has arrived; the collective can proceed.}
            \State $collectives.pop(\textit{group\_id}, \textit{call\_idx})$
            \LComment{Return all kernels to be scheduled.}
            \State \Return $\textit{wait\_list}$ 
        \Else
            \State \Comment{Not all workers have arrived; keep blocking.}
            \State \Return $\emptyset$
        \EndIf
    \EndProcedure
\EndNetworkCollectiveWaitMap
\end{algorithmic}
\end{algorithm}

%% file: tables/h100_kernel_accuracy.tex
\begin{table}[htbp]
\centering
\begin{tabular}{l l}
\toprule
\textbf{Kernel} & \textbf{MAPE}\\
\midrule
RadixSortOnesweepKernel & 7.80\% \\
cuComputeGradGammaBeta & 7.95\% \\
masked\_softmax\_warp\_backward & 0.73\% \\
compute\_num\_of\_partial\_segments & 7.37\% \\
unrolled\_elementwise\_kernel & 5.80\% \\
write\_num\_of\_segments & 7.27\% \\
cuApplyLayerNorm & 1.98\% \\
MemcpyHtoD & 14.23\% \\
CatArrayBatchedCopy\_aligned16\_contig & 5.79\% \\
cuComputeGradInput & 3.50\% \\
MemcpyDtoH & 7.85\% \\
compute\_grad\_weight & 3.63\% \\
at\_cuda\_detailcubDeviceScanKernel & 5.37\% \\
cublasSgemm\_v2 & 3.65\% \\
cublasSgemmStridedBatched & 2.22\% \\
indexSelectLargeIndex & 1.88\% \\
multi\_tensor\_apply\_kernel & 1.68\% \\
at\_cuda\_detailcubDeviceScanInitKernel & 6.99\% \\
triu\_tril\_kernel & 4.38\% \\
vectorized\_elementwise\_kernel & 8.44\% \\
krn\_partial\_segment\_offset & 55.35\% \\
RadixSortExclusiveSumKernel & 14.30\% \\
CatArrayBatchedCopy & 43.71\% \\
fused\_dropout\_kernel\_vec & 1.50\% \\
index\_elementwise\_kernel & 12.86\% \\
sum\_and\_scatter & 48.82\% \\
MemcpyDtoD & 0.00\% \\
reduce\_kernel & 16.75\% \\
RadixSortHistogramKernel & 9.01\% \\
masked\_softmax\_warp\_forward & 1.00\% \\
cuComputePartGradGammaBeta & 4.12\% \\
krn\_partials\_per\_segment & 7.16\% \\
elementwise\_kernel & 10.28\% \\
elementwise\_kernel\_with\_index & 24.67\% \\
thrustcuda\_cubcore\_kernel\_agent & 12.51\% \\
Memset & 13.25\% \\
\bottomrule
\end{tabular}
\caption{Mean absolute percentage error on a held-out validation set, trained on H100 kernel runtimes. Important kernel types for Megatron-LM models include \texttt{cublasSgemm\_v2} and \texttt{cublasSgemmStridedBatched}, where we have <5\% prediction error. Kernels with large percentage-wise errors are extremely short in duration, and thus do not impact end-to-end latency significantly.}
\label{tab:h100_kernel_accuracy}
\end{table}

%% file: tables/v100_kernel_accuracy.tex
\begin{table}[htbp]
\centering
\begin{tabular}{l l}
\toprule
\textbf{Kernel} & \textbf{MAPE}\\
\midrule
scaled\_masked\_softmax\_warp\_backward & 0.41\% \\
at\_cuda\_detailcubDeviceScanKernel & 5.80\% \\
sum\_and\_scatter & 49.87\% \\
write\_num\_of\_segments & 30.59\% \\
vectorized\_elementwise\_kernel & 11.44\% \\
indexSelectLargeIndex & 7.20\% \\
elementwise\_kernel & 26.48\% \\
krn\_partial\_segment\_offset & 48.18\% \\
fused\_dropout\_kernel\_vec & 1.03\% \\
index\_elementwise\_kernel & 10.47\% \\
cuApplyLayerNorm & 1.36\% \\
elementwise\_kernel\_with\_index & 31.91\% \\
compute\_num\_of\_partial\_segments & 10.39\% \\
cuComputePartGradGammaBeta & 3.05\% \\
MemcpyDtoH & 39.56\% \\
unrolled\_elementwise\_kernel & 13.89\% \\
Memset & 36.75\% \\
MemcpyHtoD & 25.61\% \\
CatArrayBatchedCopy & 105.45\% \\
krn\_partials\_per\_segment & 11.16\% \\
cuComputeGradInput & 1.80\% \\
cublasSgemm\_v2 & 4.58\% \\
compute\_grad\_weight & 2.23\% \\
triu\_tril\_kernel & 11.76\% \\
multi\_tensor\_apply\_kernel & 3.40\% \\
RadixSortHistogramKernel & 9.00\% \\
at\_cuda\_detailcubDeviceScanInitKernel & 14.83\% \\
masked\_softmax\_warp\_forward & 1.20\% \\
RadixSortExclusiveSumKernel & 38.65\% \\
thrustcuda\_cubcore\_kernel\_agent & 32.53\% \\
cublasSgemmStridedBatched & 1.84\% \\
cuComputeGradGammaBeta & 18.95\% \\
CatArrayBatchedCopy\_aligned16\_contig & 20.19\% \\
reduce\_kernel & 24.64\% \\
RadixSortOnesweepKernel & 13.54\% \\
MemcpyDtoD & 33.25\% \\
scaled\_masked\_softmax\_warp\_forward & 0.48\% \\
softmax\_warp\_backward & 1.04\% \\
\bottomrule
\end{tabular}
\caption{Mean absolute percentage error on a held-out validation set, trained on V100 kernel runtimes. Important kernel types for Megatron-LM models include \texttt{cublasSgemm\_v2} and \texttt{cublasSgemmStridedBatched}, where we have <5\% prediction error. Kernels with large percentage-wise errors are extremely short in duration, and thus do not impact end-to-end latency significantly.}
\label{tab:v100_kernel_accuracy}
\end{table}

%% file: tables/a40_kernel_accuracy.tex
\begin{table}[htbp]
\centering
\begin{tabular}{l l}
\toprule
\textbf{Kernel} & \textbf{MAPE}\\
\midrule
cudnnConvolutionBackwardFilter & 9.16\% \\
elementwise\_kernel & 24.35\% \\
CatArrayBatchedCopy\_aligned16\_contig & 14.96\% \\
Memset & 34.27\% \\
triton & 4.13\% \\
cudnnConvolutionBackwardData & 7.89\% \\
tensor\_kernel\_scan\_innermost\_dim & 153.91\% \\
MemcpyDtoH & 37.05\% \\
cublasSgemm\_v2 & 37.08\% \\
softmax\_warp\_forward & 229.09\% \\
MemcpyHtoD & 27.71\% \\
cudnnConvolutionForward & 6.31\% \\
multi\_tensor\_apply\_kernel & 1.51\% \\
cublasSgemmStridedBatched & 63.61\% \\
nll\_loss\_backward\_reduce\_cuda\_kernel\_2d & 253.28\% \\
softmax\_warp\_backward & 164.03\% \\
unrolled\_elementwise\_kernel & 10.98\% \\
max\_pool\_backward\_nhwc & 17.19\% \\
cublasLtMatmul & 83.92\% \\
MemcpyDtoD & 65.84\% \\
CatArrayBatchedCopy & 96.63\% \\
vectorized\_elementwise\_kernel & 18.18\% \\
distribution\_elementwise\_grid\_stride\_kernel & 228.62\% \\
nll\_loss\_forward\_reduce\_cuda\_kernel\_2d & 171.61\% \\
\bottomrule
\end{tabular}
\caption{Mean absolute percentage error on a held-out validation set, trained on A40 kernel runtimes. Important kernel types for vision models include \texttt{cudnnConvolution} and \texttt{triton}, where we have <10\% prediction error. Kernels with large percentage-wise errors are extremely short in duration, and thus do not impact end-to-end latency significantly.}
\label{tab:a40_kernel_accuracy}
\end{table}

%% file: figures-tex/search_algo_comparison.tex
\begin{figure}[h!]
    \centering
    \vspace{2em}
\includegraphics[width=0.44\textwidth,keepaspectratio]{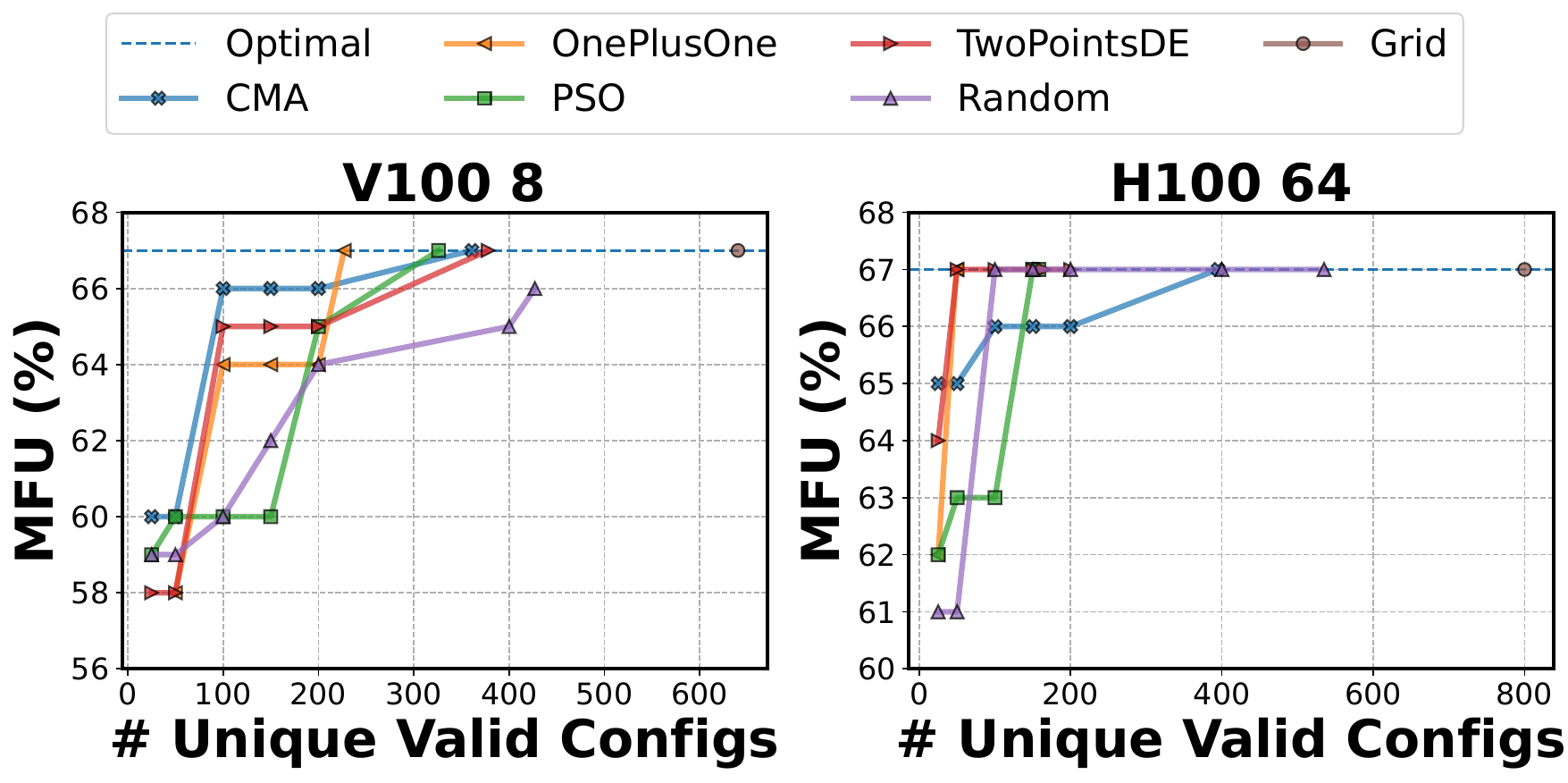}

\caption{Comparison of search algorithms exploring GPT3-2.7B (left) and GPT3-18.4B (right). Each algorithm is given a 2000 sample budget. Most algorithms achieve near-optimal MFU after 200-300 valid configurations, providing 60-75\% improvement over grid search. \amey{is the model same in both cases?}}
\label{fig:config-search-alg-comparison}
\end{figure}

%% file: tables/config_search_optimizations.tex
\begin{table*}[t]
\centering
\begin{tabular}{c p{0.3\textwidth} p{0.4\textwidth}}
\toprule
\textbf{Knob} & \textbf{Performance characteristics} & \textbf{Tactic} \\
\midrule
Activation recomputation & Reduces memory footprint through smart activation checkpointing 
 & If a prior config OOMed with activation recomputation enabled, then skip the similar config that only disables activation recomputation and mark its result as OOMed \\ 
Sequence parallelism & Reduces memory footprint by reducing activation memory with no added communication cost & If a prior config OOMed with sequence parallelism enabled, then skip the similar config that only disables sequence parallelism and mark its result as OOMed \\
Distributed optimizer & Reduces memory footprint by sharding gradient and optimizer state with added communication cost & If a prior config did not OOM without the distributed optimizer, then skip the similar config that only enables the distributed optimizer and set its runtime to be the same \\
No. of microbatches & In the absence of pipeline parallelism, hardware utilization is inversely proportional to the number of microbatches \cite{megatron}. & If a prior config did not OOM with number of microbatches $n$ and no pipeline parallelism, then skip the similar config that only increases the number of microbatches and set its runtime to be the same \\
\bottomrule
\end{tabular}
\caption{Summary of fidelity-preserving tactics used in Megatron-LM configuration search experiment}
\label{table:fidelity-preserving-tactics}
\end{table*}